\documentclass[preprint,11pt]{elsarticle}

\usepackage{graphicx}
\usepackage{amssymb}
\usepackage[left=2.5cm,right=2.5cm,top=2.5cm,bottom=2.5cm]{geometry}
\usepackage{algpseudocode, algorithm}
\usepackage{setspace}
\usepackage{subfigure}
\usepackage{tabularx}
\usepackage{array}
\usepackage{multirow}
\usepackage{amsmath,bm}
\usepackage{amsfonts}

\journal{Cognitive Systems Research}

\begin{document}
\begin{frontmatter}

\title{Neural Implementation of Probabilistic Models of Cognition}

\author{Milad Kharratzadeh$^1$, Thomas Shultz$^2$}

\address{$^{1}$Department of Electrical and Computer Engineering \\
$^{2}$Department of Psychology \& School of Computer Science \\ McGill University, Montreal, Quebec, Canada \\ milad.kharratzadeh@mail.mcgill.ca, thomas.shultz@mcgill.ca}

\begin{abstract}
Bayesian models of cognition hypothesize that human brains make sense of data by representing probability distributions and applying Bayes' rule to find the best explanation for available data. Understanding the neural mechanisms underlying probabilistic models remains important because Bayesian models  provide a computational framework, rather than specifying mechanistic processes.  Here, we propose a deterministic neural-network model which estimates and represents probability distributions from observable events --- a phenomenon related to the concept of probability matching.  Our model learns to represent probabilities without receiving any representation of them from the external world, but rather by experiencing the occurrence patterns of individual events. Our neural implementation of probability matching is paired with a neural module applying Bayes' rule, forming a comprehensive  neural scheme to simulate human Bayesian learning and inference. Our model also provides novel explanations of  base-rate neglect, a notable deviation from Bayes.
\end{abstract}

\end{frontmatter}

\section{Introduction} \label{sec:intro}
 Bayesian models are now prominent across a wide range of problems in cognitive science including inductive learning~\citep{Ten06}, language acquisition~\citep{Cha06}, and  vision~\citep{Yui06}. These models characterize a rational solution to  problems in cognition and perception in which inferences about different hypotheses are made with limited data under uncertainty. In Bayesian models, beliefs are represented by probability distributions and are updated by Bayesian inference as additional data become available. For example, the baseline probability of having cancer is lower than that of having a cold or heartburn.  Coughing is more likely caused by cancer or cold than by heartburn. Thus, the most probable diagnosis for coughing is a cold, because having a cold has a high probability both before and after the coughing is observed.   Bayesian models of cognition state that humans make inferences in a similar fashion. More formally, these models hypothesize that humans make sense of data by representing probability distributions and applying Bayes' rule to find the best explanation for available data.
 
Forming internal representations of probabilities of different hypotheses (as a measure of belief) is one of the most important components of several explanatory frameworks.  For example, in decision theory, many experiments show that participants select alternatives proportional to their frequency of occurrence. This means that in many scenarios, instead of maximizing their utility by always choosing the alternative with the higher chance of reward, they match the underlying probabilities of different alternatives. For a review,  see~\citep{Vul98}.
 
There are several challenges for Bayesian models of cognition as suggested by recent critiques~\citep{Jon11, Ebe11, Bow12, Mar13}.  First, these models mainly  operate at Marr's computational level~\citep{Mar82}, with no account of the mechanisms underlying behaviour.  That is, they are not concerned with how people actually learn and represent the underlying probabilities.   Jones and Love~(\citeyear[p. 175]{Jon11}) characterize this neglect of mechanism as ``the most radical aspect of Bayesian Fundamentalism''. Second, in current Bayesian models, it is typical for cognitive structures and hypotheses to be designed by researchers, and for Bayes' rule to select the best hypothesis or structure to explain the available evidence~\citep{Shu07}. Such models often do not typically explain or provide insight into the origin of such hypotheses and structures.  Bayesian models are under-constrained in the sense that they predict various outcomes depending on assumed priors and likelihoods~\citep{Bow12}. Finally, it is shown that people can be rather poor Bayesians and deviate from the optimal Bayes' rule due to biases such as base-rate neglect, the representativeness heuristic, and confusion about the direction of conditional probabilities~\citep{Edd82, Kah96, Ebe11, Mar13}.

In this paper, we address some of these challenges by providing a psychologically plausible neural framework to explain probabilistic models of cognition at Marr's implementation level. 
As the main component of our framework, we study how deterministic neural networks can learn to represent probability distributions; these distributions can serve later as priors or likelihoods in a Bayesian framework. 
 We consider deterministic networks because from a modelling perspective, it is important to see whether randomness and probabilistic representations can emerge as a property of a population of deterministic units rather than a built-in property of individual stochastic units. For our framework to be psychologically plausible it requires two important properties: (i) it needs to learn the underlying distributions from observable inputs (e.g., binary inputs indicating whether an event occurred or not) and (ii) it needs to adapt to the complexity of the distributions or changes in the probabilities. We discuss these aspects in more details later.

The question of how people  perform Bayesian computations (including probability representations) can be answered at two levels~\citep{Mar82}. First, it can be explained at the level of psychological processes, showing that Bayesian computations can be carried out by modules similar to the ones used in other psychological process models~\citep{Kru06}. Second, probabilistic computations can  also be treated at a neural level, explaining how these computations could be performed by a population of connected neurons~\citep{Ma06}. Our artificial neural network framework combines these two approaches. It provides a neurally--based model of probabilistic learning and inference that can be used to simulate and explain a variety of psychological phenomena.

 We use this comprehensive modular neural implementation of Bayesian learning and inference to explain some of the well-known deviations from Bayes' rule, such as base-rate neglect, in a neurally plausible fashion. In sum, by providing a psychologically plausible implementation-level explanation of probabilistic models of cognition, we integrate some seemingly contradictory accounts within a unified framework.

The paper is organized as follows. First, we review  necessary background material and introduce the problem's setup and notation. Then, we introduce our proposed framework for realizing probability matching with neural networks. Next, we present empirical results and discuss some relevant phenomena often observed in human and animal learning. Finally, we propose a neural implementation of Bayesian learning and inference, and show that  base-rate neglect can be implemented by a weight-disruption mechanism.

\section{Learning Probability Distributions via Deterministic Units}\label{sec:NNPT}
\subsection{Problem Setup}
\begin{figure}[!b]
\centering
\includegraphics[scale=0.4]{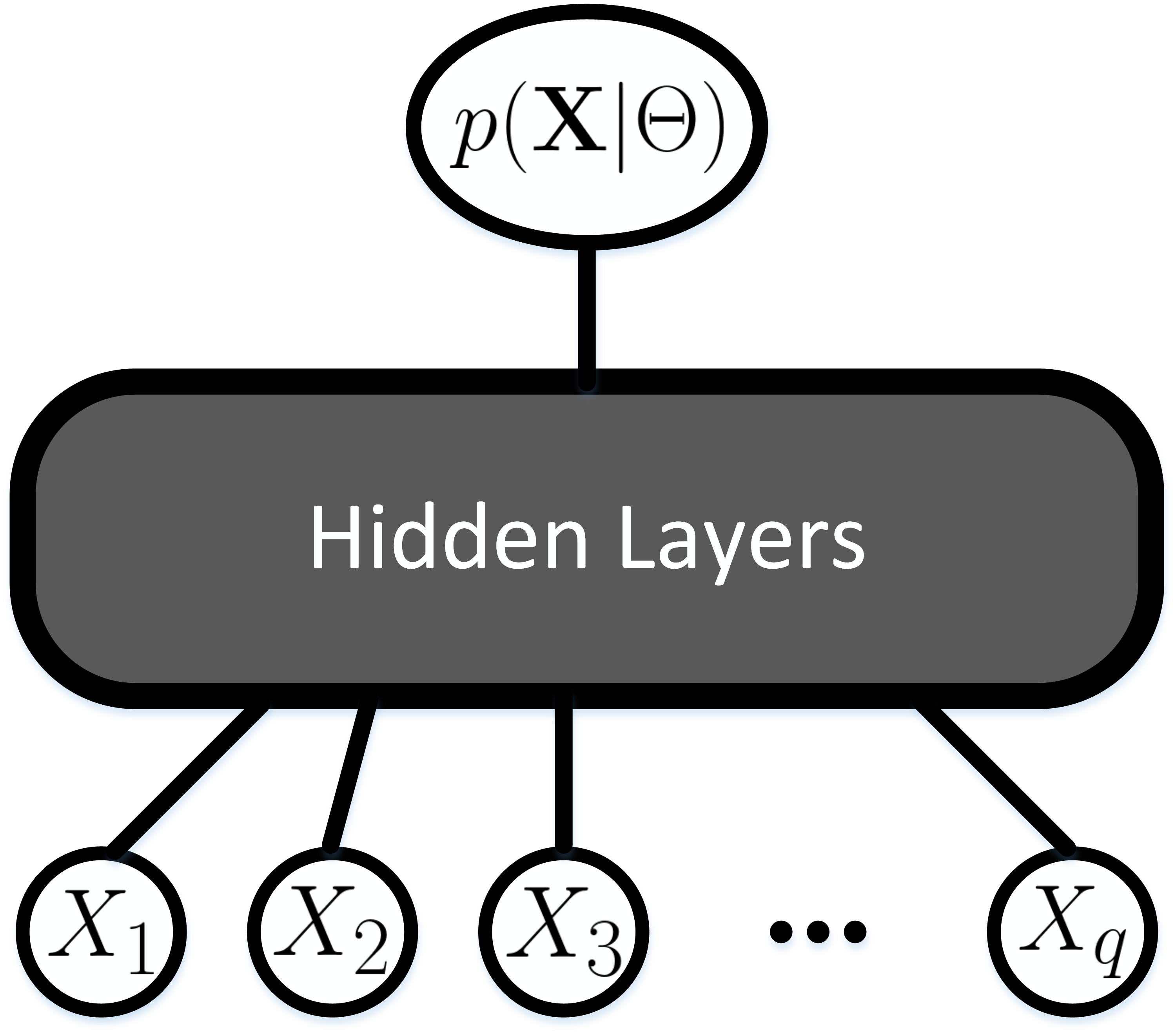}
\caption{The basic structure of the network learning a $q$-dimensional probability distributions. Both  structural details and connection weights in the hidden layers are learned.\label{fig:nnmodel}}
\end{figure}

The first goal of this paper is to introduce networks that learn probability distributions from realistic inputs. We consider the general case of multivariate probability distributions defined over $q\geq1$ different random variables, ${\bf X} = (X_1,X_2,\ldots,X_q)$. We represent the value of the density function by $p({\bf X}|\Theta)$, where $\Theta$ represents the functional form and parameters of the distribution. We assume that $\Theta$ is unknown in advance and thus would need to be learned. As shown in Fig.~\ref{fig:nnmodel}, the neural network learning this multivariate distribution has $q$ input units corresponding to $(X_1,X_2,\ldots,X_q)$ and one output unit corresponding to $p({\bf X}|\Theta)$.  

\paragraph*{\bf Realistic Inputs} In real-world scenarios, observations are in the form of events which can occur or not (represented by outputs of 1 and 0, respectively) under various conditions and the learner does not have access to the actual probabilities  of those events. The most important requirement for our framework is the ability to learn using realistic patterns corresponding to occurrences or non-occurrences of events in different circumstances. For instance, consider a simplified example where $q=2$, $X_1$ is the month of the year and $X_2$ is the weather type (rainy, sunny, snowy, stormy, etc.), and $p(X_1,X_2)$ represents their joint distribution. An observer makes different observations over the years; for instance on a rainy August day, the observations are  $(X_1=\text{August}, X_2=\text{rainy}, {\bf 1})$, $(X_1=\text{August}, X_2=\text{sunny}, {\bf 0})$, $(X_1=\text{August}, X_2=\text{snowy}, {\bf 0})$, etc., where $1$ denotes the occurrence of an event and $0$ denotes the non-occurrence. Over the years, and after making many observations of this type, people form an internal approximation of $p(X_1,X_2)$. We assume that the training sets for our networks are similar: each training sample is a realization of the input vector $(X_1=x_1, \ldots, X_q=x_q)$ paired with a binary $0$ or $1$ in the output unit. Note that, the target output for input $(X_1=x_1, \ldots, X_q=x_q)$ is $p(X_1=x_1, \ldots, X_q=x_q)$, but because these probabilities are rarely available in real world, the outputs in the training set are binary and the network has to learn the underlying probability distribution from these binary observations. 

\paragraph*{\bf Adaptiveness and Autonomous Learning} As noted, we assume no prior information about the form of the underlying distribution; it can be a simple uniform distribution or a complicated multi-modal one. A psychologically plausible framework must learn autonomously; it should start with small computational power (e.g., a few hidden units) and increase the complexity of the network structure (e.g., by adding more hidden layers) until it learns the underlying distribution successfully. Moreover, the network should be able to detect and quickly adapt to the changes in the underlying distribution. 

In the next section, we propose a learning framework that satisfies all these conditions. In the remainder of this section, we review   work that studied related problems and discuss our differences.

\subsection{Related Work}
Our proposed scheme differs from the classical approach to neural networks in a number of ways. First, in our framework, there is no one-to-one relationship between inputs and output. Instead of being paired with one fixed output, each input is here paired with a series of $1$s and $0$s presented separately at the output unit. Moreover, in our framework, the underlying probabilities are hidden from the network  and, in the training phase, the network is presented only with inputs and their probabilistically varying binary outputs.   

The relationship between neural network learning and probabilistic inference has been studied previously. One approach is to use networks with stochastic units that fire with particular probabilities. Boltzmann machines~\citep{Ack85} and their various derivatives, including Deep Learning in hierarchical restricted Boltzmann machines (RBM)~\citep{Hin06}, have been proposed to learn a probability distribution over a set of inputs. RBM  tries to maximize the likelihood of the data using a particular graphical model. In an approach similar to Boltzmann machines, Movellan and McClelland introduced a class of stochastic networks called Symmetric Diffusion Networks (SDN) to reproduce an entire probability distribution (rather than a point estimate of the expected value) on the output layer~\citep{Mov93}. In their  model, unit activations are probabilistic functions evolving from a system of stochastic differential equations. ~\cite{Mcc98} showed that a network of stochastic units can estimate likelihoods and posteriors and make ``quasi-optimal'' probabilistic inference. 
Sigmoid type belief networks, a class of neural networks with stochastic units, providing a framework for representing probabilistic information in a variety of unsupervised and supervised learning problems, perform Bayesian calculations in a computationally efficient fashion~\citep{Jord1,Jord2}. 
A multinomial interactive activation and competition (mIAC) network, which has stochastic units, can correctly sample from the posterior distribution and thus, implement optimal Bayesian inference~\citep{Mcc14}. However, the presented mIAC model is specially designed for a restricted version of the word recognition problem and is highly engineered due to  preset biases and weights and preset organization of units into multiple pools.

Instead of assuming stochastic units, we show how probabilistic representations can be constructed by the output of a population of deterministic units.  These deterministic units fire at a rate which is a sigmoid function of their net input. Moreover, models with stochastic units such as RBM ``require a certain amount of practical experience to decide how to set the values of numerical meta-parameters''~\citep{Hin10}, which makes them neurally and psychologically implausible to model the relatively autonomous learning of humans or animals. On the other hand, as we see later, our model learns the underlying distributions in a relatively autonomous, neurally--plausible fashion, by using deterministic units in a constructive learning algorithm that builds the network topology as it learns.  

Probabilistic interpretations of deterministic  back--propagation (BP) learning have also been studied~\citep{Rum95}. Under certain restrictions, BP can be viewed as learning to produce the most likely output, given a particular input. To achieve this goal, different cost functions (for BP to minimize) are introduced for different distributions~\citep{Mcc98}. This limits the plausibility of this model in realistic scenarios, where the underlying distribution might not be known in advance, and hence the appropriate cost function for BP cannot be chosen a priori. Moreover, the ability to learn probabilistic observations has been shown only  for members of the exponential family where the distribution has that specific form. In contrast, our model is not restricted to any particular type of probability distribution, and there is no need to adjust the cost function to the underlying distribution in advance. Also, unlike BP, where the structure of the network is fixed in advance, our constructive network learns both weights and the structure of the network in a more autonomous fashion, resulting in a psychologically plausible model.

Neural networks with simple, specific structures have been proposed for specific probabilistic tasks~\citep{Shan90, Shan91, Lop98, Daw09, Grif12, Mcc14}.  For instance, \citep{Grif12} considered a specific model of property induction and observed that, for certain distributions, a linear neural network shows a similar performance to Bayesian inference with a particular prior.  Dawson et al. proposed a neural network to learn probabilities for a multiarm bandit problem~\citep{Daw09}. The structure of these neural networks were engineered to suit the problem being learned. In contrast, our model is general in that it can learn probabilities for any problem structure. Also, unlike previous models proposing neural networks to estimate the posterior probabilities~\citep{Ham90}, our model does not require explicit representations of the probabilities as inputs. Instead, it constructs an internal representation based on observed patterns of occurrence.

\section{The Learning Algorithm}
In this section, we study the feasibility of a learning framework with all the constraints described in the last section. First, we show that it is theoretically possible to learn the underlying distribution from realistic inputs. Then, we propose an algorithm based on sibling-descendant cascade correlation and learning cessation techniques that can  learn the probabilities in an autonomous and adaptive fashion. 
 
\subsection{Theoretical Analysis}\label{sec:theory}
The statistical properties of feed-forward neural networks with deterministic units have been studied as non-parametric density estimators. Consider a general case where both the the input,${\bf X}$, and output,${\bf Y}$, of a network can be multi-dimensional. In a probabilistic setting, the relationship between $\bf X$ and $\bf Y$ is determined by the conditional probability $p({\bf Y}|{\bf X})$. White (\citeyear{Whi89}) and Geman et al. (\citeyear{Gem92}) showed that under certain assumptions, feed-forward neural networks with a single hidden layer can consistently learn the conditional expectation function $E({\bf Y}|{\bf X})$. However, as White mentions, his analyses ``do not provide more than very general guidance on how this can be done'' and suggest that ``such learning will be hard''~\citep[p. 454]{Whi89}. Moreover, these analyses ``say nothing about how to determine adequate network complexity in any specific application with a given training set of size $n$''~\citep[p. 455]{Whi89}. In our work, we first consider a more general case with no restrictive assumptions about the structure of the network and learning algorithm. Then, we propose a learning algorithm that automatically determines the adequate network complexity in any particular application.

As shown in Fig.~\ref{fig:nnmodel}, our model has a single output, and  we have $Y\in\{0,1\}$ which indicates whether an event occurred or not. In this case $E(Y=1|{\bf X}) = p(Y=1|{\bf X})$. Thus, successful learning is equivalent to representing the underlying probabilities in the output unit. 

{\bf Theorem.} Assume that we have a multivariate distribution, $p({\bf X})$, and $N$ training samples of the form $({\bf x}_i, y_i)$, where $y_i=1$ with probability $p({\bf x}_i)$ and $y_i=0$ otherwise. Define the network error as the sum--of--squared error at the output:

\begin{equation} \label{eq:sos}
\mathcal{E} = \frac{1}{2} \sum_{i=1}^N (o_i - y_i)^2.
\end{equation}
where $o_i$ is the network's output when ${\bf x}_i$ is presented at the input. Then, any learning algorithm that successfully trains the network to minimize the output sum--of--squared error results in learning the distribution $p$, i.e., for any input ${\bf x}'$, the output will be $p({\bf x}')$. 

{\bf Proof.} Denote the output of the network for the input ${\bf x}'$ by $o$. When the error is minimized, we have 
\begin{align}
\frac{\partial \mathcal{E}}{\partial o} = 0 \Rightarrow \sum_{i: \ {\bf x}_i = {\bf x}'} (o - y_i) = 0 \Rightarrow o = {\sum_{i: \ {\bf x}_i = {\bf x}'} y_i}/{N'},
\end{align}
where $N'$ is the number of times ${\bf x}'$ appeared as the input. Therefore, as $N'\to\infty$, according to the strong law of large numbers $o \overset{a.s.}{\to} p({\bf x}')$, where $\overset{a.s.}{\to}$ denotes almost sure convergence. Therefore, the network's output converges to the underlying probability distribution, $p$, at all points. 

This theorem makes the important point that neural networks with deterministic units are able to asymptotically estimate an underlying probability distribution solely based on observable binary outputs. Unlike previous similar results in literature~\citep{Whi89, Gem92, Rum95}, our  theorem does not impose any constraint on the network structure, the learning algorithm, or the distribution being learned. However, an important assumption in this theorem is the successful minimization of the error by the learning algorithm. Two important questions remain to be answered: (i) how can this learning be done? and (ii) how  can adequate network complexity be automatically identified for a given training set? In the next two subsections, we address these two problems and propose a learning framework to successfully minimize the output error.

\subsection{Learning Cessation}\label{sec:cess}
In artificial neural networks, learning normally continues until an error metric is less than a fixed small threshold. However, that approach may lead to overfitting and also would not work here, because the least possible error is a positive constant instead of zero. 
We use the idea of learning cessation to overcome these limitations~\citep{Shu12n}. The learning cessation method monitors learning progress in order to autonomously abandon unproductive learning. It checks the absolute difference of consecutive errors and if this value is less than a fixed threshold multiplied by the current error for a fixed number of consecutive learning phases (called patience), learning is abandoned. This technique for stopping deterministic learning of stochastic patterns does not require the psychologically unrealistic validation set of training patterns~\citep{Pre98, Wan93}. 

Our method (along with the learning cessation mechanism) is presented in Algorithm~1. In this algorithm, we represent the whole network (units and connections) by the variable $Net$. Also, the learning algorithm we use to train our network is represented by the operator {\it train\_one\_epoch}, where an epoch is a pass through all of the training patterns. We can use any algorithm to train our network, as long as it satisfies the conditions mentioned in the problem setup and successfully minimizes the error term in (\ref{eq:sos}). We discuss the details of the learning algorithm in the next subsection.

\begin{algorithm}[!h]
\caption{Probability matching with neural networks and learning cessation}
\begin{spacing}{1.1}
\begin{algorithmic}
\State \textbf{Input:} Training Set $S_{train} = \{ (h_i, r_{ij}) \ | \ h_i \in X \ ; r_{ij} \sim \textrm{Bernoulli}(P(h_i)) \}$;  
\State \qquad \quad Cessation threshold $\epsilon_c$; Cessation patience $patience$
\State \textbf{Output:} Learned network outputs $\{ o_i \ , \ i=1,\ldots,m \}$
\State $counter \gets 0, t \gets 0$
\While {true}
\State $(\{ o_i \ | \ i=1,\ldots,m \}, Net) \gets \textrm{train\_one\_epoch} (Net, S_{train})$ \Comment{Updating the network}
\State $\mathcal{E}_p(t) \gets  \frac{1}{2} \sum_{i=1}^m \sum_{j=1}^n (o_i - r_{ij})^2$ \Comment{Computing the updated error}
\If {$ \ |\mathcal{E}_p(t)-\mathcal{E}_p(t-1)|   \geq \epsilon_c \cdot |\mathcal{E}_p(t)| \ $} \Comment{Checking the learning  progress}
    \State {$counter \gets 0$}
\Else
\State $counter \gets counter + 1$
\If {$ \ counter = patience $}
\State \textbf{break}
\EndIf
\EndIf
\State $t \gets t+1$
\EndWhile
\end{algorithmic}
\end{spacing}
\end{algorithm}

\subsection{Autonomous Learning via a Constructive Algorithm}\label{sec:lal}
We showed that the minimization of the output sum--of--squared error is equivalent to learning the probabilities. However, the realistic training set we employ as well as the fact that we do not know the functional form or parameters of the underlying distribution in advance may cause problems for some neural learning algorithms. The most widely used learning algorithm for neural networks is Back Propagation, also used by Dawson et al., (2009) in the context of learning probability distributions. In Back Propagation (BP), the output error is propagated backward and the connection weights are individually adjusted to minimize this error.  Despite its many successes in cognitive modelling, we do not recommend using BP in our scheme for two important reasons. First, when using BP, the network's structure must be fixed in advance (mainly heuristically). This makes it impossible for the learning algorithm to automatically adjust the network complexity to the problem at hand (White, 1989). Moreover, this property limits the generalizability and autonomy of BP and also,  along with back-propagation of error signals, makes it psychologically implausible. Second, due to their fixed design, BP networks are not suitable for cases where the  underlying distribution changes over time. For instance, if the distribution over the hypotheses space gets  more complicated over time, the initial network's complexity (i.e., number of hidden units) would fall short of the required computational power. In sum, BP fails the autonomy and adaptiveness conditions we require in our framework.

 Instead of BP, we use a variant of the cascade correlation (CC) method called sibling-descendant cascade correlation (SDCC) which is a constructive method for learning in multi-layer artificial neural \mbox{networks~\citep{Bal94}}. SDCC learns both the network's structure and the connection weights; it starts with a minimal network, then automatically trains new hidden units and adds them to the active network, one at a time. Each new unit is employed at the current or a new highest layer and is the best of several candidates at tracking current network error.

 The SDCC network starts with a perceptron topology, with input units coding the example input and output units coding the correct response to that input (see Fig.~\ref{fig:nnmodel}). In constructive fashion, neuronal units are recruited into the network one at a time as needed to reduce error. In classical CC, each new recruit is installed on its own layer, higher than previous layers. The SDCC variant is more flexible in that a recruit can be installed either on the current highest layer (as a sibling) or on its own higher layer as a descendant, depending on which location yields the higher correlation between candidate unit activation and current network error~\citep{Bal94}. In both CC and SDCC, learning progresses in a recurring sequence of two phases -- output phase and input phase. In output phase, network error at the output units is minimized by adjusting connection weights  without changing the current topology. In the input phase, a new unit is recruited such that the correlation between its activation and network error is maximized. In both phases, the optimization is done by the Quickprop algorithm~\citep{Fahl98}.

 SDCC offers two major advantages over BP. First, it constructs the network in an autonomous fashion (i.e., a user does not have to design the topology of the network, and also the network can adapt to environmental changes). Second, its greedy learning mechanism can be orders of magnitude faster than the standard BP algorithm~\citep{Fah90}.
SDCC's relative autonomy in learning  is similar to humans' developmental, autonomous learning~\citep{Shu12-2}. With SDCC, our method implements psychologically realistic learning of probability distributions, without any preset topological design. The psychological and neurological validity of cascade-correlation and SDCC has been well documented in many publications~\citep{Shu13, Shu13-2}. These algorithms have been shown to accurately simulate a wide variety of psychological phenomena in learning and psychological development. Like all useful computational models of learning, they abstract away from neurological details, many of which are still unknown. Among the principled similarities with known brain functions, SDCC exhibits distributed representation, activation modulation via integration of neural inputs, an S-shaped activation function, layered hierarchical  topologies, both cascaded and direct pathways, long-term potentiation, self-organization of network topology, pruning, growth at the newer end of the network via synaptogenesis or neurogenesis, weight freezing, and no need to back-propagate error signals. 


\subsection{Sampling from a Learned Distribution}
\begin{figure}
\centering
\includegraphics[scale=0.27]{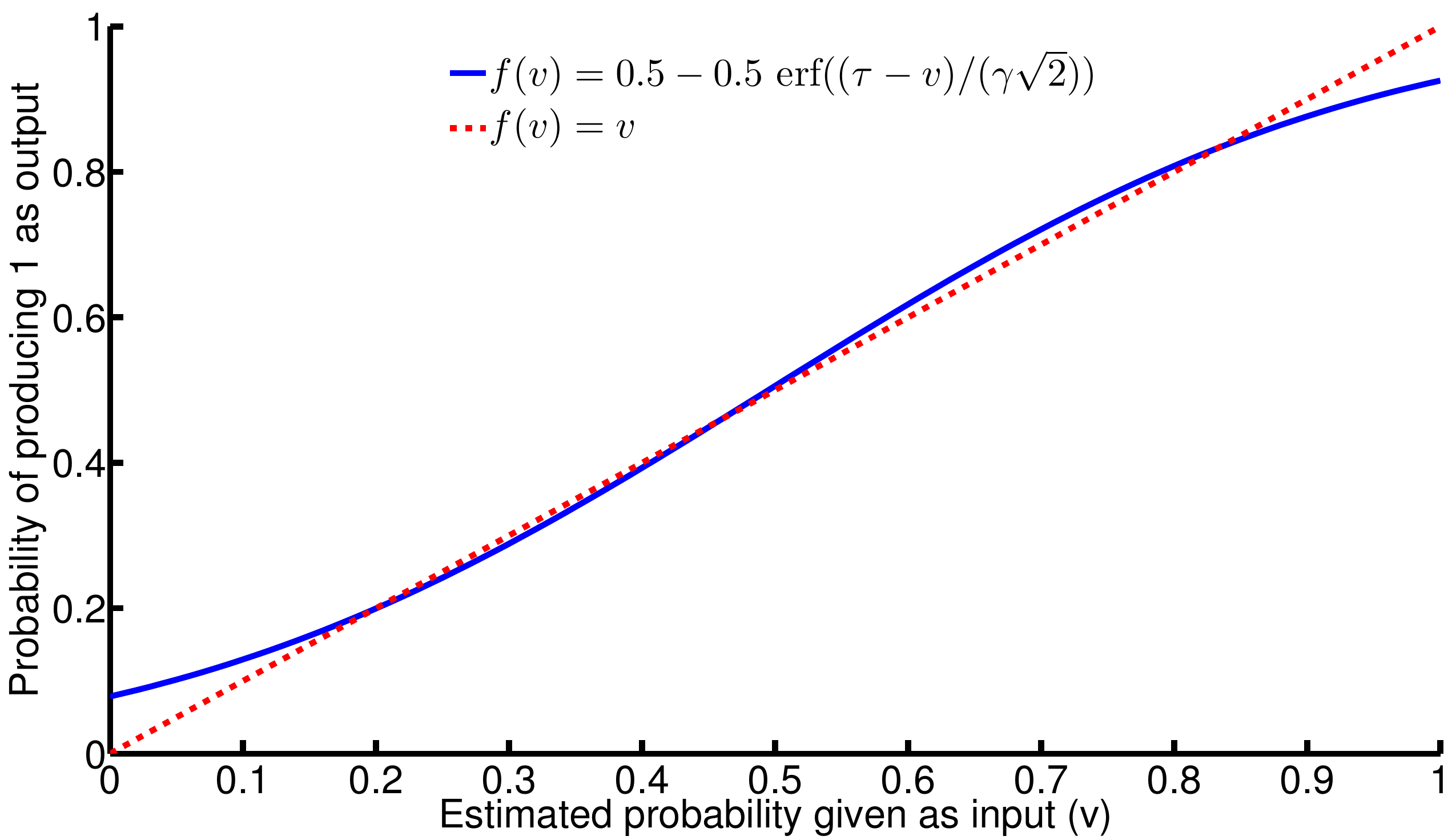}
\caption{A deterministic unit with a thresholding activation function generating responses that match the probabilities that each response is correct ($\tau=1, \gamma=0.35$)\label{fig:erf}}
\end{figure}

 So far, we have described a framework to learn multivariate distributions using deterministic units. It is shown, e.g., in the context of probability matching~\citep{Vul98}, that people are able to choose alternatives proportional to their underlying probabilities. In this part, we discuss how our networks' estimated probabilities can be used with deterministic units to produce binary samples from $p(X_1, X_2, \ldots, X_q)$, i.e., for input $(x_1,\ldots,x_q)$ generating 1 in the output with probability $p(x_1, \ldots, x_q)$ and 0 otherwise. We show that deterministic units with simple thresholding activation functions and added Gaussian noise in the input can generate probabilistic samples. Assume that we have a neuron with two inputs: the estimated probability produced by our network for a certain input configuration, $0\leq v\leq 1$, and a zero--mean Gaussian noise, $\epsilon \sim \mathcal{N}(0,\gamma)$. Then, given the thresholding activation function, the output will be 1 if $v+\epsilon>\tau$ and 0 if $v+\epsilon\leq\tau$ for a given threshold~$\tau$. Therefore, the probability of producing 1 at the output is:

\begin{equation}
p(\textrm{output} = 1 | v,\tau,\gamma) = p(v+\epsilon>\tau) = p(\epsilon>\tau-v) = \underbrace{0.5 - 0.5 \  \textrm{erf}\left(\frac{\tau-v}{\gamma\sqrt{2}}\right)}_{f(v)},
\end{equation}
where {\it erf} denotes the error function: $\operatorname{erf}(x) = (2/\sqrt\pi)\int_0^x e^{-t^2}\,\mathrm dt$. It is easy to see that $f(v)$ lies between 0 and 1 and, for appropriate choices of $\tau$ and $\gamma$, we have $f(v)\simeq v$ for $0<v<1$ (see Fig.~\ref{fig:erf}). Thus, a single thresholding unit with additive Gausian noise in the input can use the estimated probabilities to produce responses that approximate the trained response probabilities. This allows sampling from the learned distribution. 

\section{Simulation Results} \label{sec:simul}
In this section, we provide  simulation results to study the properties of our proposed learning framework. We examine the accuracy, scalability and adaptiveness of learning.

\subsection{Accuracy}
\begin{figure}[!h]
     \begin{center}
        \subfigure[Discrete distribution]{%
            \label{fig:pm_dis}
            \includegraphics[angle=270, width=0.3\textwidth]{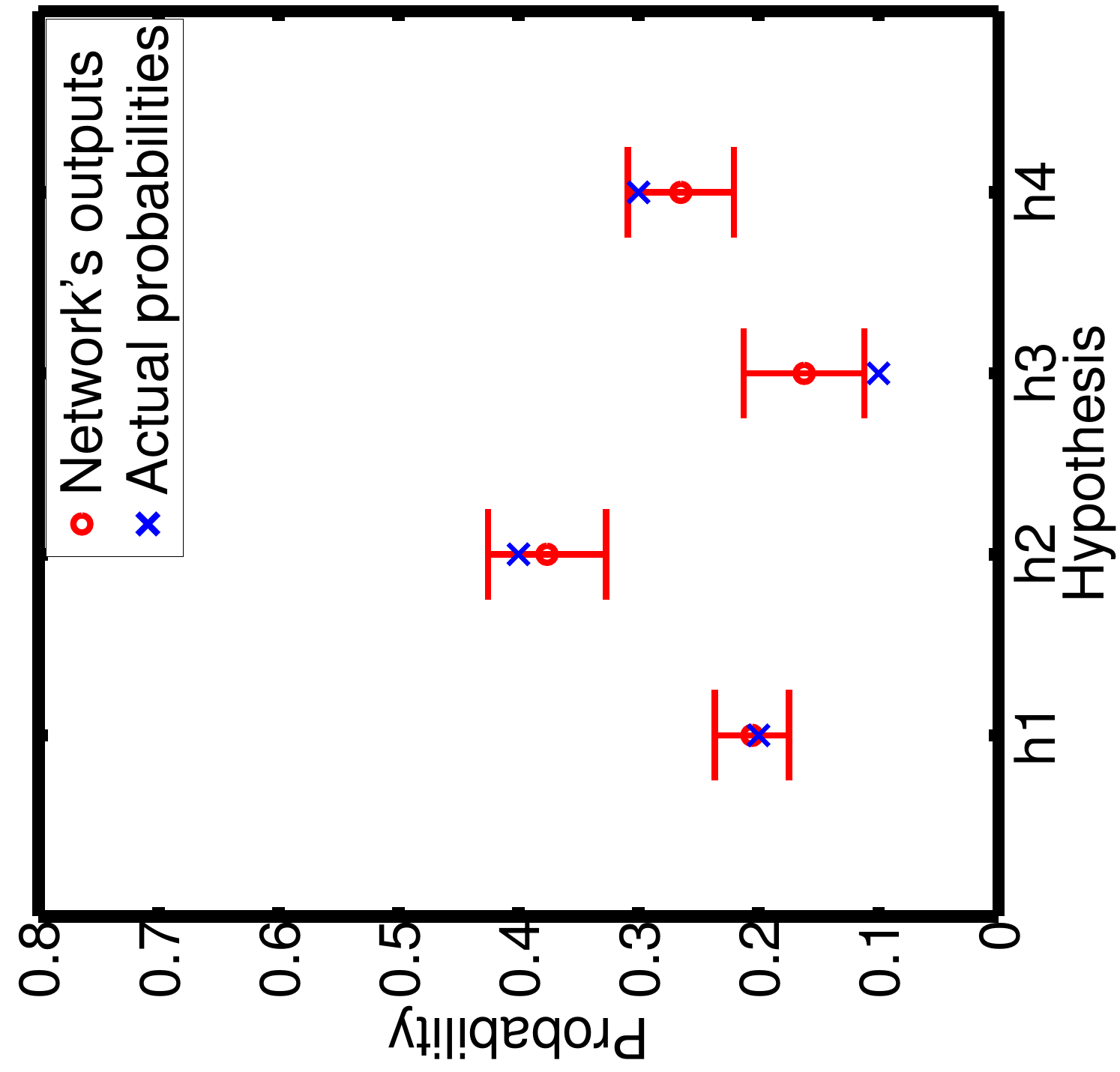}
        }%
        \subfigure[Continuous distribution (Normal)]{%
           \label{fig:pm_con}
           \includegraphics[angle=270, width=0.53\textwidth]{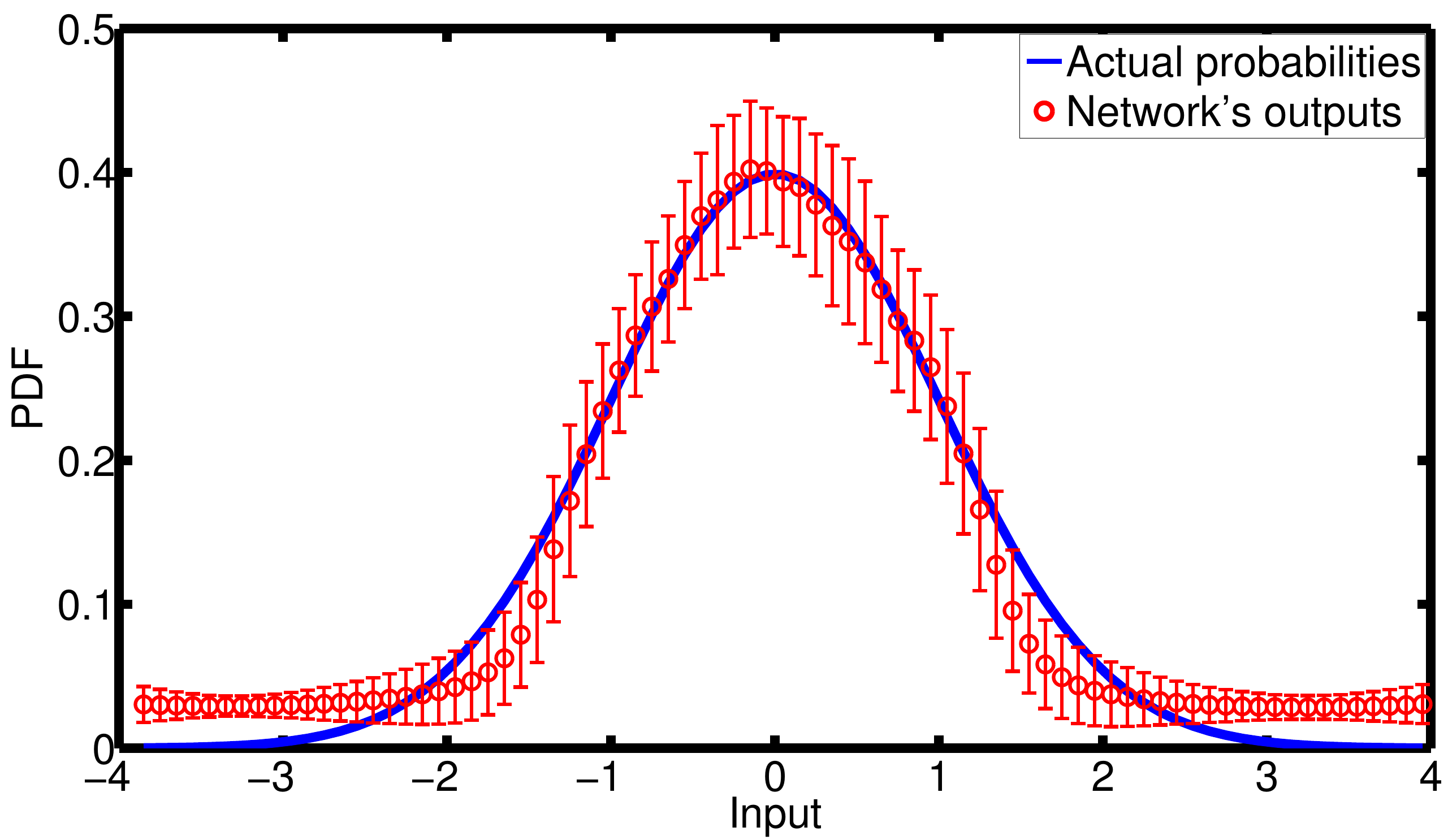}
        } 
    \end{center}
    \caption{%
    Learning of the underlying probability distribution by our SDCC model. The results (mean and standard deviation) are averaged over $50$ different networks.       
     }%
   \label{fig:pm}
\end{figure}

We first examine whether  our proposed scheme is successful in learning the underlying distributions.  We start with one-dimensional distributions and consider two cases here, but we observed similar results for a wide range of probability distributions. First, we consider a case of four hypotheses with probability values $.2, .4, .1, $ and $.3$. Also, we consider a Normal probability distribution where the hypotheses correspond to small intervals on the real line from $-4$ to 4.  For each input sample we consider $15$ randomly selected instances in each training epoch. As  before, an output event occurs with a target probability. We use SDCC with learning cessation to train our networks. Fig.~\ref{fig:pm}, plotted as the average and standard deviation of the results for $50$ networks, demonstrates that for both discrete and continuous probability distributions, the network outputs are close to the actual distribution. Although, to save space here, we show the results for only two sample distributions, our experiments show that this  model is able to learn a wide range of one-dimensional distributions including Binomial, Poisson, Gaussian, and Gamma~\citep{Kha13}.

\begin{figure}[!h]
\centering
\includegraphics[scale=0.24]{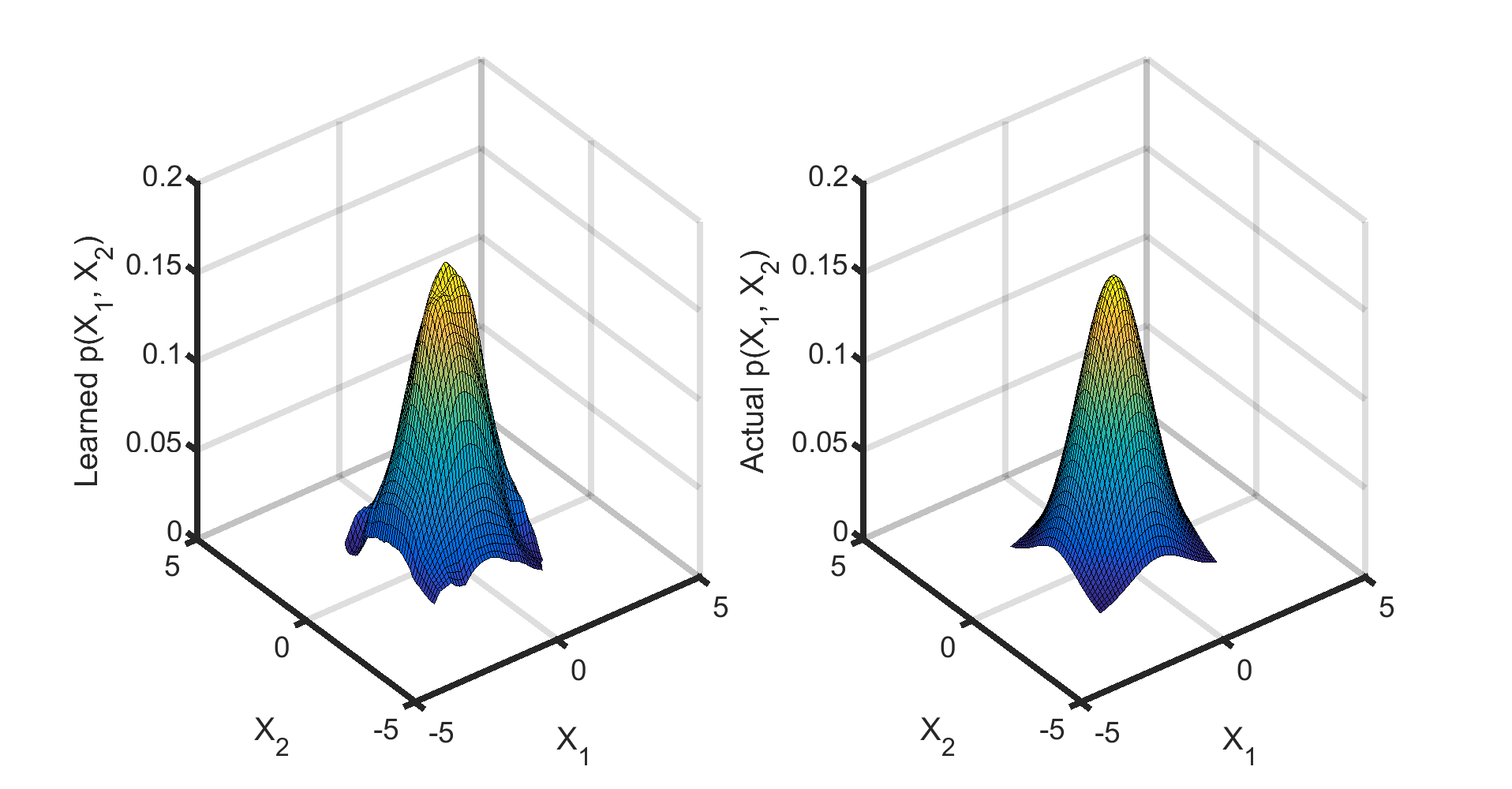}
\caption{Learning the underlying distribution for a two-dimensional Gaussian distribution. The left plot shows the outputs of our network for a test set which has not been seen during the training. \label{fig:G2D2}}
\end{figure}

We can perform the same study for multivariate distributions. In Fig.~\ref{fig:G2D2}, we show the actual distribution function of a 2D Gaussian with mean $(0,0)$ and identity covariance matrix on the right, and the output average of 50 networks learned by our algorithm on the left. We observe that our scheme is successful in learning the underlying distribution. The learning is done over a lattice: $X_1$ and $X_2$ vary from $-2$ to 2 in steps of size $0.1$. It is important to note that for assessing the generalization accuracy of our networks, in Fig.~\ref{fig:G2D2} (and also Fig.~\ref{fig:pm_con}), we plot the output of our network for a test set which has not been seen during the training. Input variables $X_1$ and $X_2$ vary from $-2.05$ to $2.05$ in steps of size $0.1$. 

It is important to study the scalability of the learned networks when the size of the training samples or the dimensionality of the data increases. To examine this, we continue with the Gaussian distribution with zero mean and identity covariance matrix. In the training set, each of the random variables, $X_i$, varies from $-2$ to 2 in steps of size $0.2$ (i.e., total of 21 samples per dimension). Therefore, as we increase the dimension of the Gaussian distribution from 1 to 4, the size of the sample space increases from 21 to $21^4$. For each problem, we continue the training until the same level of accuracy is achieved (i.e., correlation between the true values and network outputs more than $0.7$). We observe that for these sample spaces, the median size of the network (over 50 runs) changes from $6$ units for the 1-dimension problem to $16$ units for the 4-dimension problem. As we go to even higher dimensions, the prohibitive factor would be the sample size. For instance, for a 10-dimensional input, the sample size in this example will be $21^{10}$ (roughly 
$1.4E13$) which is too large for training. However, this happens if we want to learn the distribution over the whole 10-dimensional domain. Normally, it is sufficient to learn the distribution only over a small subspace of the domain, in which case the learning will be feasible again.

\subsection{Adaptiveness}
\begin{figure}[ht!]
     \begin{center}
 \hspace*{-0.2 in}        \subfigure[Discrete case]{%
            \label{fig:bchng}
            \includegraphics[angle=270, width=0.48\textwidth]{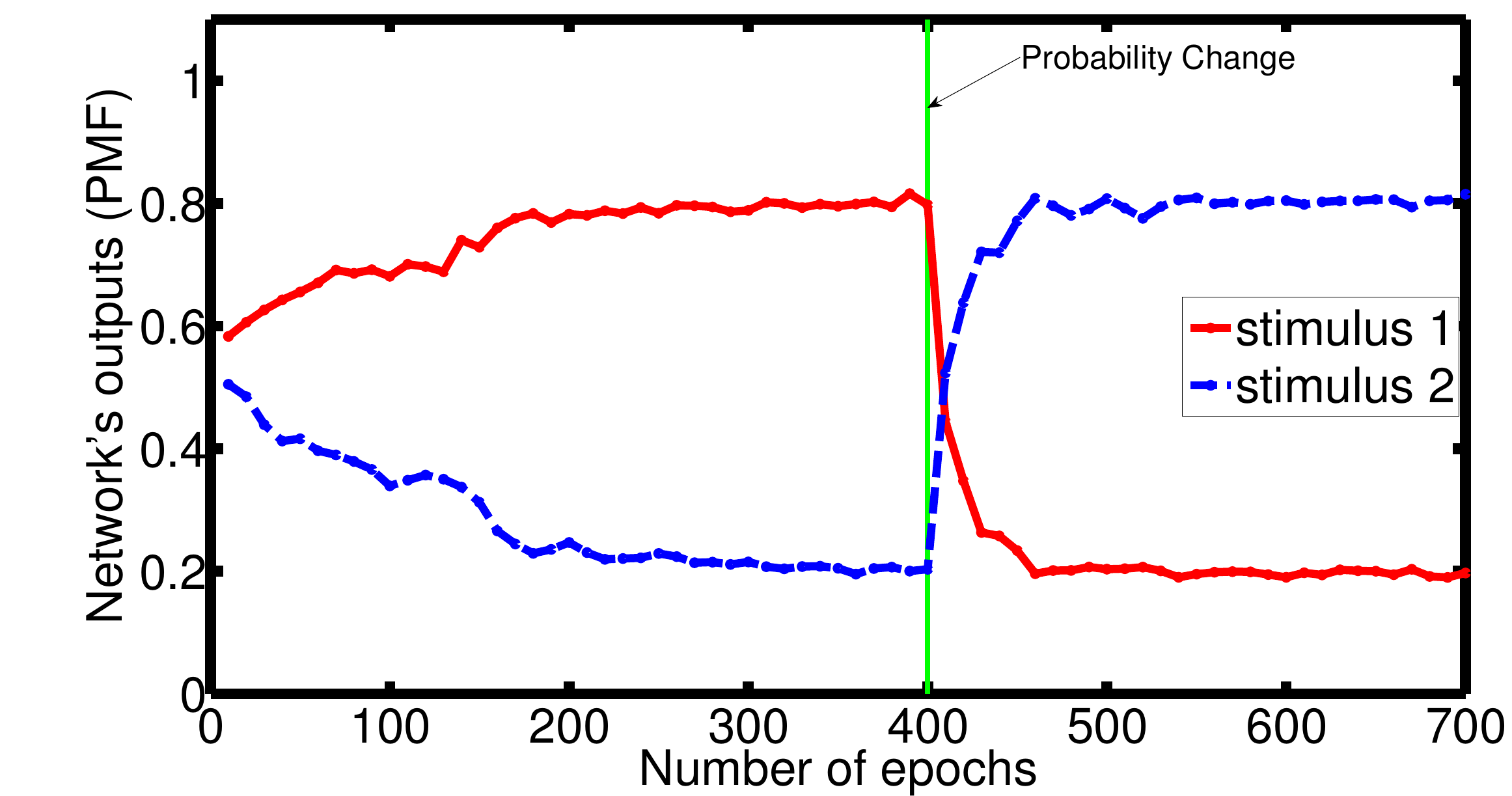}
        }%
        \subfigure[Continuous case]{%
           \label{fig:gchng} 
           \includegraphics[angle=270, width=0.46\textwidth]{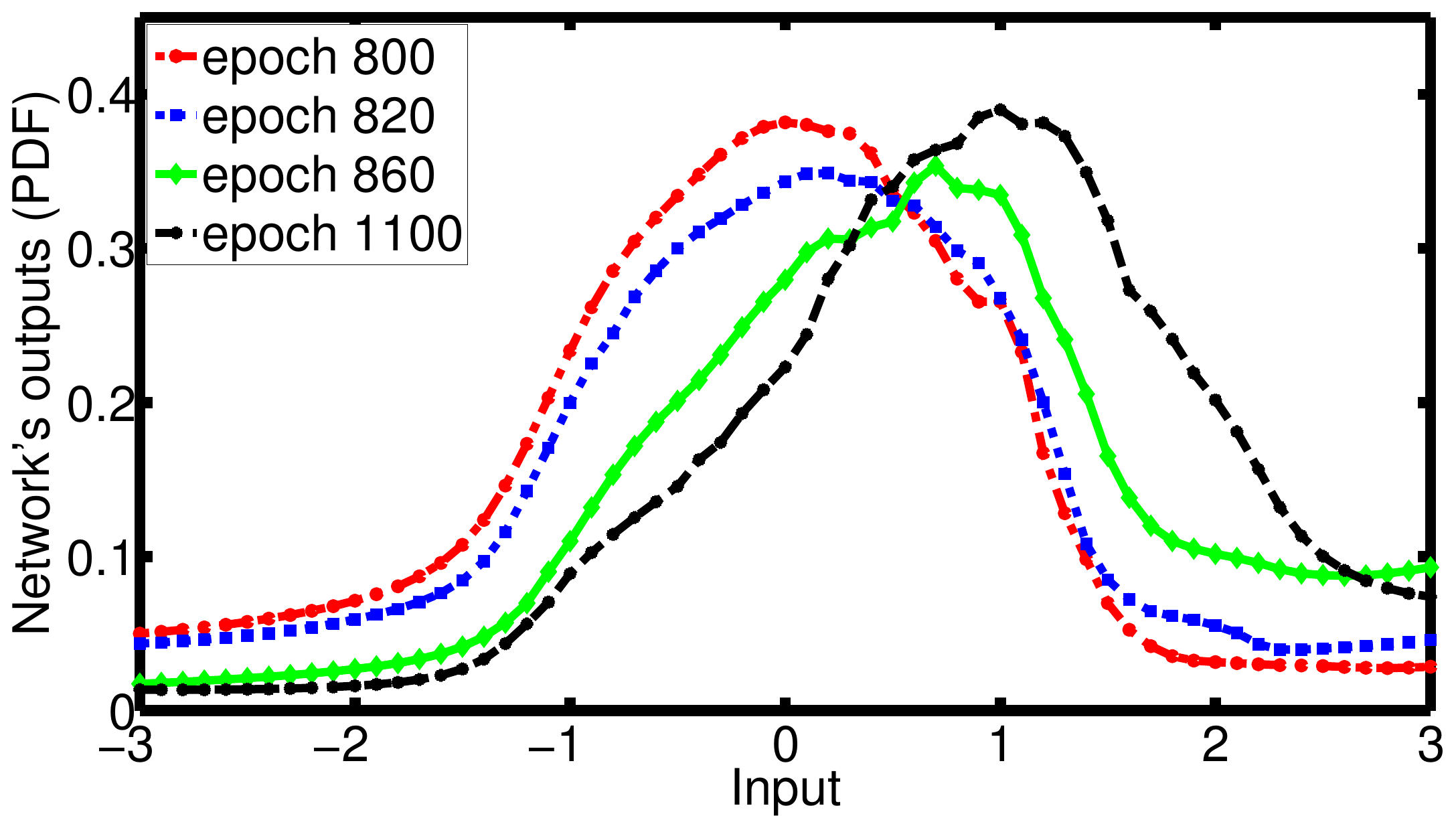}
        }%
    \end{center}
    \caption{%
       Reaction of the network to the changes in target probabilities. Our networks can adapt successfully.
     }%
   \label{fig:bayes}
\end{figure}

In many natural environments, the underlying reward patterns change over time. For example, in a Bayesian context, the likelihood of an event can change as the underlying conditions change. Because humans are able to adapt to such changes and update their internal representations of probabilities, successful models should have this property as well. We examine this property in the following example experiment. Assume we have a binary distribution where the possible outcomes have probabilities $.2$ and $.8$, and these probabilities change after $400$ epochs to $.8$ and $.2$, respectively. In Fig.~\ref{fig:bchng}, we show the network's outputs for this scenario.  We perform a similar simulation for the continuous case where the underlying distribution is Gaussian and we change the mean from $0$ to $1$ at epoch $800$; the network's outputs are shown in Fig.~\ref{fig:gchng}. We observe that in both cases, the network successfully updates and matches the new probabilities.

We also observe that adapting to the changes takes less time than the initial learning. For example, in the discrete case, it takes 400 epochs to learn the initial probabilities while it takes around 70 epochs to adapt to the new probabilities. The reason is that for the initial learning,  constructive learning has to grow the network until it is complex enough to represent the probability distribution. However, once the environment changes, the network has  enough computational capability to quickly adapt to the environmental changes with only a few internal changes (in weights and/or structure). We verify this in our experiments. For instance, in the Gaussian example, we observe that all 20 networks recruited 5 hidden units before the change and 11 of these networks recruited  1 and 9 networks recruited 2 hidden units afterwards.  We know of no precise psychological evidence for this reduction in learning time, but our results serve as a prediction that could be tested with biological learners. This would seem to be an example of the beneficial effects of relevant existing knowledge on new learning.

\section{Probability Matching} 

So far, we have shown that our neural-network framework is capable of learning the underlying distributions of a sequence of observations. This learning of probability distributions is closely related to the phenomenon of probability matching.
 The matching law states that the rate of a response is proportional to its rate of observed reinforcement and 
has been applied to many problems in psychology and economics~\citep{Her61, Her00}. A closely related empirical phenomenon is probability matching where the predictive probability of an event is matched with the underlying probability of its outcome~\citep{Vul98}. This is in contrast with the reward-maximizing strategy of always choosing the most probable outcome. The apparently suboptimal behaviour of probability matching is a long-standing puzzle in the study of decision making under uncertainty and has been studied extensively. 

There are numerous, and sometimes contradictory, attempts to explain this choice anomaly. Some suggest that probability matching is a cognitive shortcut driven by cognitive limitations~\citep{Vul98, Wes03}. Others assume that matching is the outcome of misperceived randomness which leads to searching for systematic patterns even in random sequences~\citep{Wol04, Wol00}. It is shown that as long as people do not believe in the randomness of a sequence, they try to discover regularities in it to improve accuracy~\citep{Unt07, Yel69}.  It is also shown that some of those who perform  probability matching in random settings have a higher chance of finding a pattern in non-random settings~\citep{Gai08}.  In contrast to this line of work, some researchers argue that probability matching reflects a mistaken intuition and can be overridden by deliberate consideration of alternative choice strategies~\citep{Koe09}. James and Koehler (\citeyear{Jam11}) suggest that a sequence-wide expectation regarding aggregate outcomes might be a source of the intuitive appeal of matching. It is also shown that people adopt an optimal response strategy if provided with (i) large financial incentives, (ii) meaningful and regular feedback, or (iii) extensive training~\citep{Sha02}.

We believe that our neural-network framework is compatible with all these accounts of probability matching. Firstly, probability matching is the norm in both humans~\citep{Woz10} and animals~\citep{Beh61, Kir65, Gre93}. It is clear that in these settings agents who match probabilities  form an internal representation of the outcome probabilities. Even for particular circumstances where a maximizing strategy is prominent~\citep{Gai08, Sha02}, it is necessary to have some knowledge of the distribution in order to produce optimal-point responses. Having a sense of the distribution provides the flexibility to  focus on the most probable point (maximizing), sample in proportion to probabilities (matching), or even generate expectations regarding aggregate outcomes (expectation generation),  all of which are evident in psychology experiments.

\section{Bayesian Learning and Inference}

\subsection{The Basics}
The Bayesian framework addresses the problem of updating beliefs in a hypothesis in light of observed data, enabling new inferences. Denote the observed data by $d$ and assume we have a set of mutually exclusive and exhaustive hypotheses, $\mathcal{H}=\{h_1, \ldots, h_N\}$, and want to infer which of these hypotheses best explains observed data (both the observations and hypotheses spaces can be multi-dimensional). In the Bayesian setting, the degrees of belief in different hypotheses are represented by probabilities. A simple formula known as Bayes' rule governs Bayesian inference. This rule specifies how the posterior probability of a hypothesis (the probability that the hypothesis is true given the observed data) can be computed using the product of data likelihood and prior probabilities:

\begin{equation} \label{bayesrule}
p(h_i|d)  =  \frac{p(d|h_i) p(h_i)}{P(d)} =  \frac{p(d|h_i) p(h_i)}{\sum_{i=1}^{N} p(d|h_i) p(h_i)}. 
\end{equation}

The probability with which we would expect to observe the data if a hypothesis were true is specified by likelihoods, $p(d|h_i)$. Priors, $p(h_i)$, represent our degree of belief in a hypothesis before observing data.  The denominator in~(\ref{bayesrule}) is called the marginal probability of data and is a normalizing sum which ensures that the posteriors for all hypotheses sum to 1.

In the Bayesian framework, we assume there is an underlying mechanism to generate the observed data. The role of inference is to evaluate various hypotheses about this mechanism and choose the  most likely mechanism responsible for generating the data. In this setting, the generative processes are specified by probabilistic models (i.e., probability densities or mass functions). 

\subsection{Max--Product Modular Network for Bayesian Inference}
Bayesian models of cognition hypothesize that human brains make sense of data by representing probability distributions and applying Bayes' rule to find the best explanation for any given data. One of the main challenges for Bayesian modellers is to explain how these two tasks (representing probabilities and applying Bayes' rule) are implemented in the brain's neural  circuitry~\citep{Per11}. We have addressed the first task (learning and representing probabilities) so far and showed that it can be implemented by our autonomous and adaptive framework. In this section, we explain how these learned probabilities can be used for efficient Bayesian inference.

We propose a modular max-product network for maximum a posteriori (MAP) inference. The comprising modules of this max-product network are SDCC networks that learn probabilities as described in previous sections.  These modules correspond to prior and likelihood distributions. For an inference problem over $N$ possible hypotheses, we need $N+1$ different modules: $N$ modules for learning the likelihood distributions for each hypothesis, $p(d|h_i)$, and one module for learning the prior distribution over all hypotheses.  

The modules are learned from realistic training samples in the form of patterns of events. For instance, in a coin flip example, assume $h_1$ is the hypothesis that a typical coin is fair. A person has seen a lot of coins and observed the results of flipping them. Because most coins are fair, hypothesis $h_1$ is positively reinforced most of the times in those experiences (and very rarely negatively reinforced).  Therefore, based on the binary feedback on the fairness of coins, our prior module  forms a high prior (close to 1) for hypothesis $h_1$. This is in accordance with the human assumption (prior) that a typical coin is most probably fair. Likelihood representations could be formed in a similar fashion based on binary feedback; in the coin example, $h_1$ is reinforced if the number of observed heads and tails in small batches of coin flips (available in  short-term memory) are approximately equal and negatively reinforced otherwise.

\begin{figure}[!b]
\centering
\includegraphics[scale=0.57]{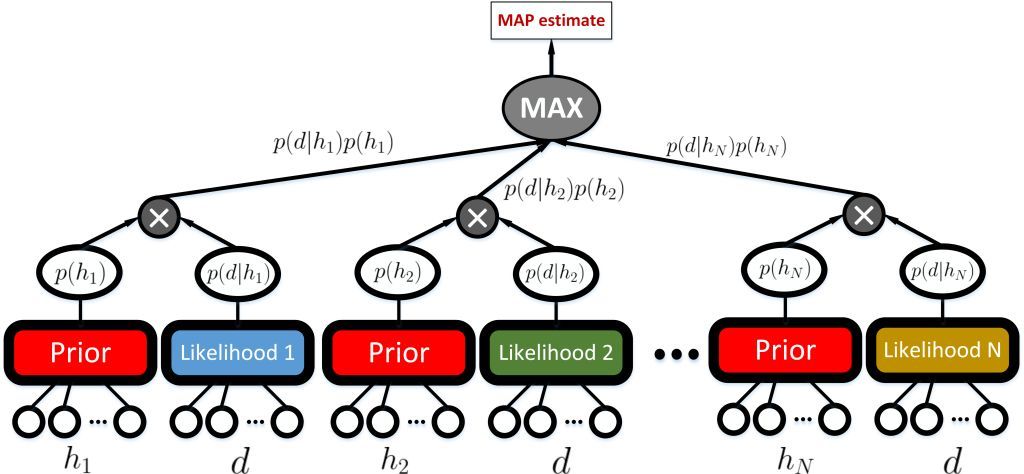}
\caption{Modular max-product network for MAP inference.\label{fig:BayesInf}}
\end{figure}

Our modular max-product network is shown in Fig.~\ref{fig:BayesInf}. The MAP inference is carried out by calculating the products $p(d|h_i)p(h_i)$ and choosing the hypothesis which gives the highest result. The normalizing term, often intractable, is common for all hypotheses and thus we can ignore it. Because of the parallel structure of our max-product network, MAP inference can be done very fast and efficiently. Also, for a given hypothesis or observation, the approximation to possibly complicated and intractable distributions can be computed efficiently by our neural modules. 

The max-product network introduced here for implementing Bayesian inference has two important benefits. First, it is an initial step towards addressing the implementation of  Bayesian competencies in the brain. Our model is built in a constructive and autonomous  fashion in accordance with  accounts of psychological development~\citep{Shu12-2}. It uses realistic training samples in the form of patterns of events and it successfully explains some phenomena often observed in human and animal learning (e.g., probability matching and adapting to environmental changes), and it can perform inference in an efficient, parallel fashion. 

The second benefit of our modular  network is that it provides a framework that unifies the Bayesian accounts and some of the well-known deviations from it, such as base-rate neglect. In the next subsection, we show how base-rate neglect can be explained naturally as a property of our neural implementation of Bayesian inference.

\begin{figure}[!h]
     \begin{center}
   \subfigure[Network structure]{%
            \label{fig:bm1}
    \includegraphics[width=0.4\textwidth]{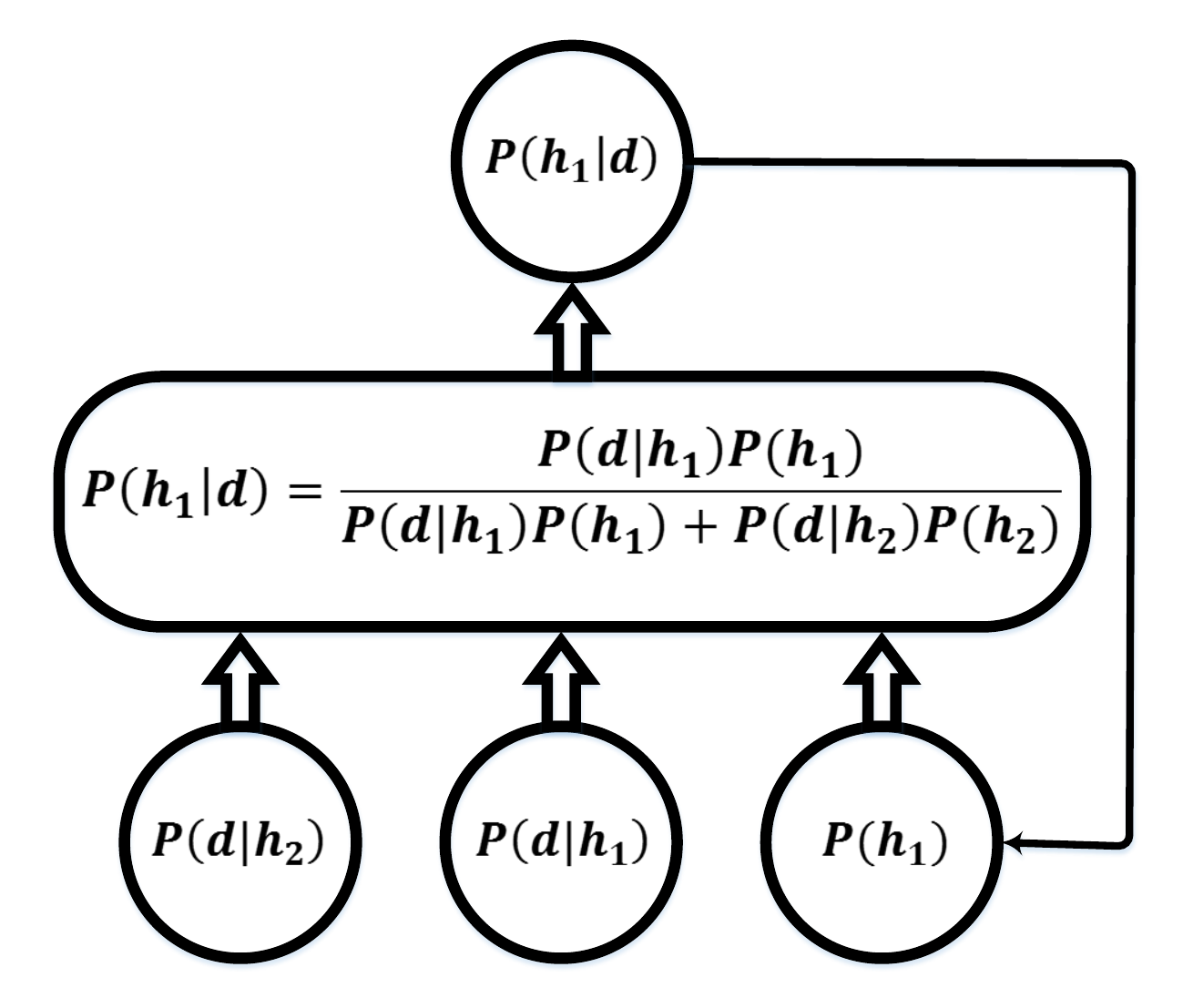}
        }%
        \subfigure[Outputs of the Bayes' rule module plotted against true values.]{%
           \label{fig:bm2} 
        \includegraphics[width=0.5\textwidth]{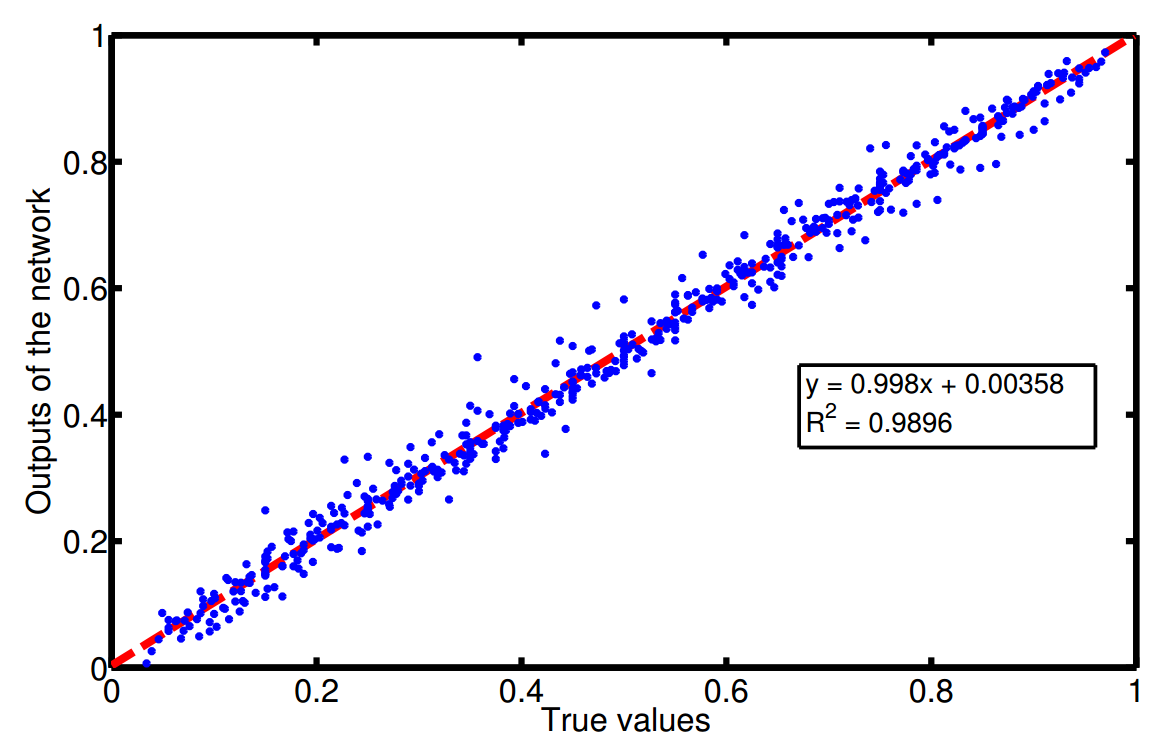}
        }%
    \end{center}
    \caption{Calculating  posterior probabilities by learning  Bayes' rule with a separate module.} \label{fig:BayesModule}
\end{figure}

\paragraph*{Exact Posterior Computation} Our max-product network performs a fast MAP inference for the general case. It is also possible to compute the exact posteriors using SDCC networks. For instance, if we have two hypotheses, the exact Bayesian inference can be carried out as depicted in Fig.~\ref{fig:bm1}, where  Bayes' rule is learned by a separate module. The outputs of the distribution modules are the inputs to  the Bayes' rule module which in turn produces the posterior probabilities on its output. In  sequential inference, this posterior can be used as a prior for the next round. In Fig.~\ref{fig:bm2}, we show that an SDCC network is successful in learning  Bayes' rule.

\subsection{Base-rate Neglect as Weight Disruption}

Given likelihood and prior distributions, the Bayesian framework finds the precise form of the posterior distribution, and uses that to make inferences. This is used in contemporary cognitive science to define rationality in learning and inference, where it is frequently defined and measured in terms of conformity to Bayes' rule~\citep{Ten06}. However, this appears to conflict with the Nobel-prize-winning work showing that people are somewhat poor Bayesians due to biases such as base-rate neglect, representativeness heuristic, and confusing the direction of conditional probabilities~\citep{Kah96}. For example, by not considering priors (such as the frequency of a disease), even experienced medical professionals deviate from optimal Bayesian inference and make major errors in their probabilistic reasoning~\citep{Edd82}. More recently, it has been suggested that base rates (i.e., priors) may not be entirely ignored but just de-emphasized~\citep{Pri11, Eva02}.  

In this section, we show that base-rate neglect can be explained in our neural implementation of the Bayesian framework. First, we show how  base-rate neglect can be interpreted by Bayes' rule. Then we show that this neglect can result from  neurally--plausible weight disruption in a neural network representing priors.

\paragraph*{\bf De-emphasizing Priors}
Base-rate neglect is a Bayesian error in computing the posterior probability of a hypothesis without taking full account of the priors. We argue that completely ignoring the priors is equivalent to assigning equal prior probabilities to all the hypotheses which gives:

\begin{equation}
P(h_i|d) = \frac{P(d|h_i)}{\sum_{i=1}^{N} P(d|h_i)} \label{eq:base-rate}.
\end{equation}
This equation can be interpreted as follows. We can assume that in the original Bayes' rule, all the hypotheses have equal priors and these priors are cancelled out from the numerator and denominator to give equation~(\ref{eq:base-rate}). Therefore, in the Bayesian framework, complete base-rate neglect is translated into assuming equal priors (i.e., equi-probable hypotheses). This means that the more the true prior probabilities (base rates) are averaged out and approach the uniform distribution, the more they are neglected in Bayesian inference. A more formal way to explain this phenomenon is by using the notion of entropy, defined in information theory as a measure of uncertainty. Given a discrete hypotheses space $\{h_1,\ldots,h_N\}$ with probability mass function $p(\cdot)$, its entropy is defined as:
$\textrm{Entropy}(X) = -\sum_{i=1}^N p(h_i)\log_2 p(h_i)$.
Entropy quantifies the expected value of information contained in a distribution. It is easy to show that a uniform distribution has maximum entropy among all discrete distributions over the hypotheses set~\citep{Cov06}. We can conclude that in the Bayesian framework, base-rate neglect is equivalent to ignoring the priors in the form of averaging them out to get a uniform distribution, or equivalently, maximizing their entropy. 

\paragraph*{\bf Weight Disruption}
In neural networks, weight disruption is a natural way to model a wide range of cognitive phenomena including the effects of attention, memory indexing, and relevance. In particular, a neural weight disruption mechanism provides a unifying framework to cover several causes of neglecting base rates: immediate effects such as deliberate neglect (as being judged irrelevant)~\citep{Bar80}, failure to recall, partial use or partial neglect, preference for specific (likelihood) information over general (prior) information~\citep{Mccl85}, and decline in some cognitive functions (such as memory loss) as a result of long term synaptic decay or interference~\citep{Har13}. 

In our model, we implement this weight disruption with the help of an {\bf attention module} which applies specific weight factors to the various modules. This weight--disruption factor reflects the strength of memory indexing or lack of relevance in a specific instance of inference, without permanently affecting the weights. It could also simulate long--term synaptic decay or interference which creates  more permanent weight disruption in a neural network. The attention module multiplies all the connection weights of a module by an attention parameter ratio, $r$, between $0$ and $1$. (Note that the disruption is applied to the connections in the network and not directly to the output.) For $r=1$, the weights of a module remain unchanged, while $r=0$ sets all the weights to zero, causing a flat output. The attention module affects both prior and likelihood modules; however, since likelihoods are formed based on recent evidence ($d$), the attention parameter for likelihood modules could be set close to $1$. For a prior module, we could allocate an attention factor $0<r<1$ to reflect partial neglect (e.g., to model partial recall, preference for specific information, or partial synaptic decay or interference) or set $r=0$ for complete neglect (e.g., to model failure to recall, deliberate neglect, or long-term synaptic decay).  


Mathematically, after a probability module is learned, its network's connection weights are updated as follows:

\begin{equation} \label{wdecay}
W_{new} = r^{t} \ W_{old}
\end{equation}
where $W$'s are the connection weights, $r \in (0,1)$ is the attention factor imposed by the attention module, and $t\in\{1, 2, 3, \dots\}$ is the number of times the factor $r$ is applied. For instantaneous disruptions, such as the cases where a prior network is not recalled or is judged irrelevant, $t=1$ and $r$ is a low number, considerably less than 1. For long--term decay, $r$ would be slightly less than 1, while $t$ would be large (modelling slow synaptic decay over long time). For higher values of $t$ and lower values of $r$, the weight disruption is more severe; with $r=1$, the weights remain unchanged, while with $r=0$, they are set to zero. 

\begin{figure}[!b]
     \begin{center}
   \subfigure[The entropy of prior distributions increases and gets closer to the uniform as disruption gets larger.]{%
            \label{fig:entropy}
    \includegraphics[width=0.4\textwidth]{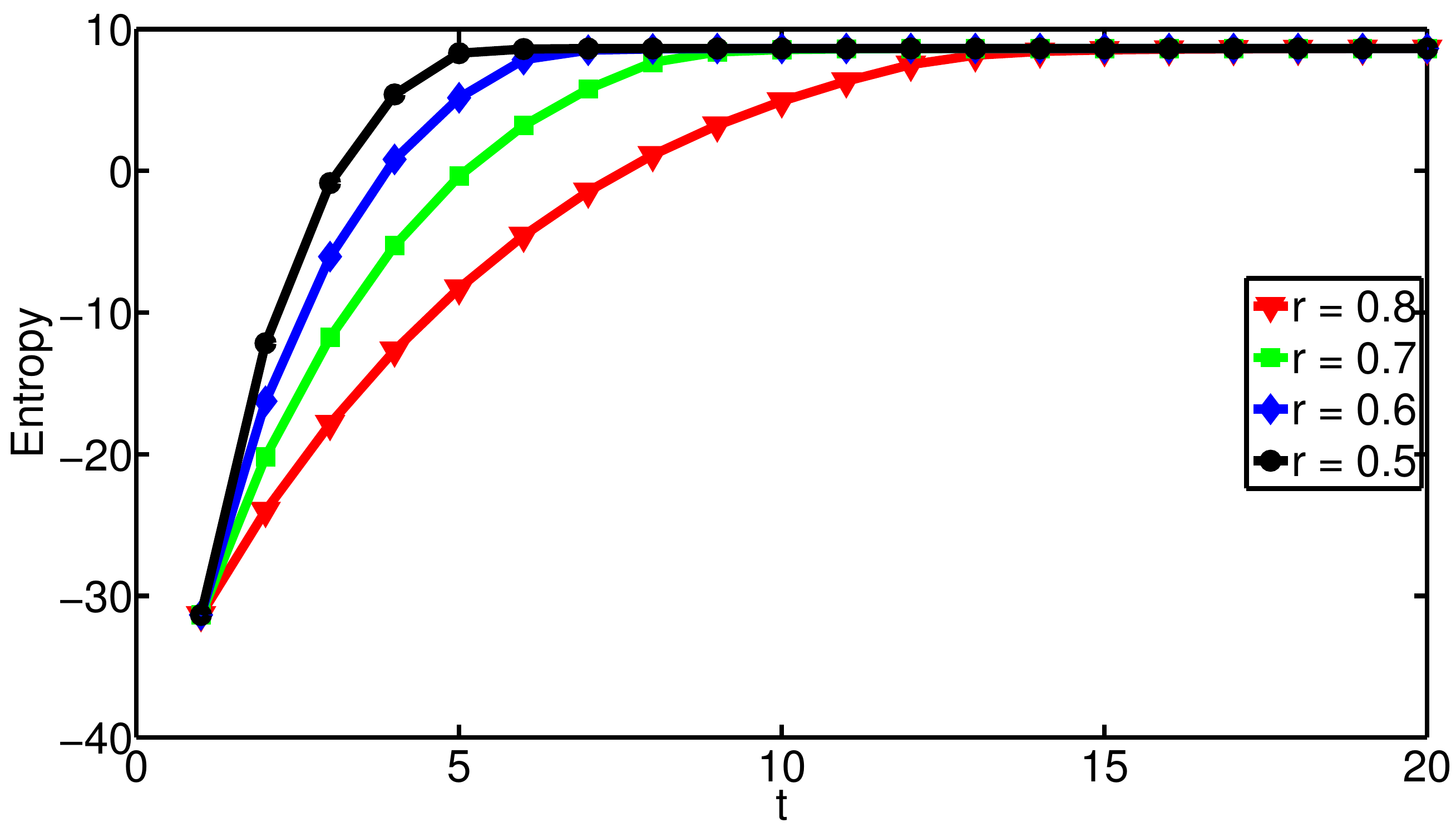}
        }~ %
        \subfigure[The distribution of the priors approaches uniform as disruption increases.]{%
           \label{fig:unif} 
        \includegraphics[width=0.4\textwidth]{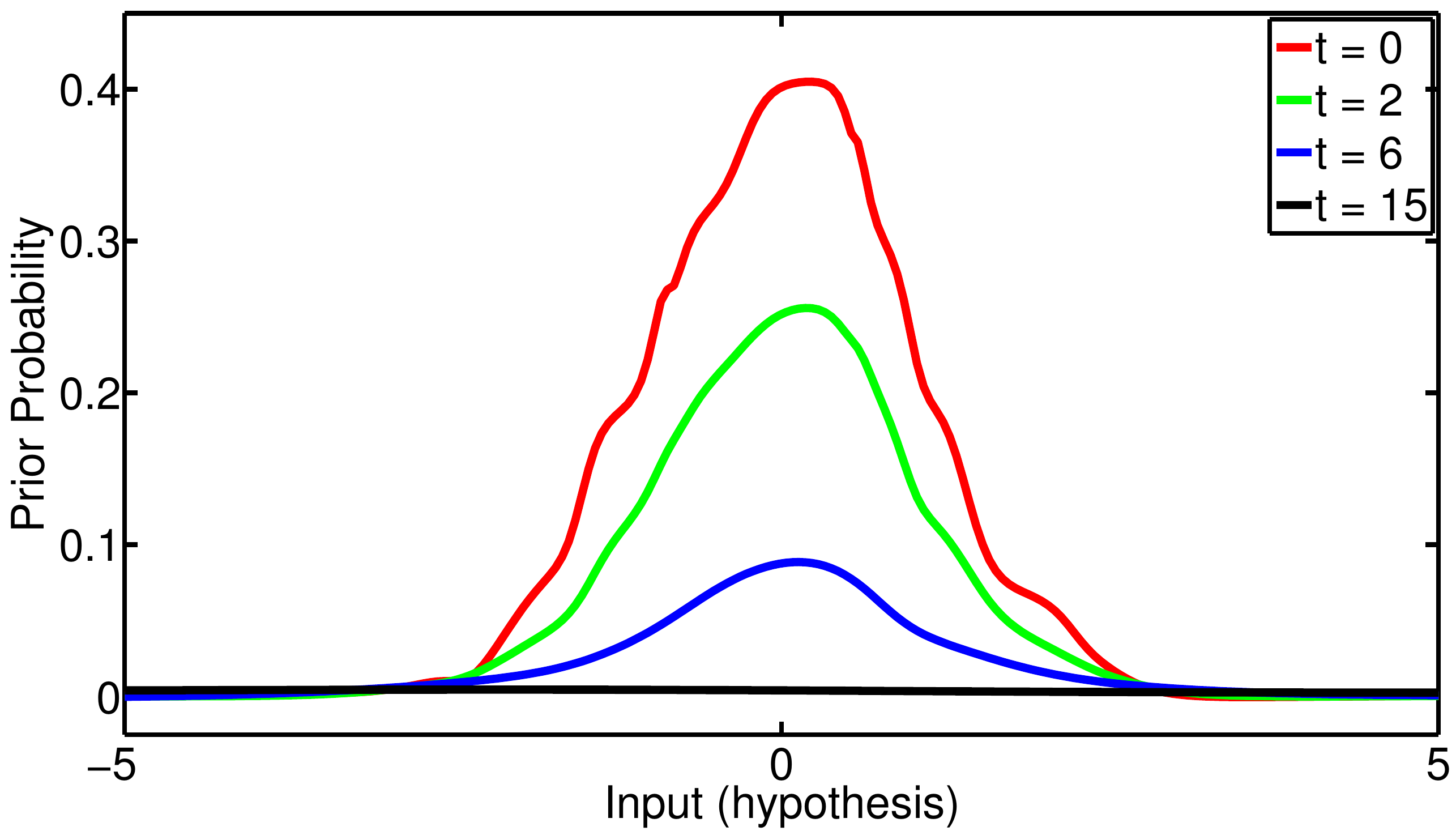}
        }%
    \end{center}
    \caption{The effects of weight disruption on the output of probability matching module.} \label{fig:wdecay}
\end{figure}

We examine the effects of our proposed weight disruption with a set of simulations. Results for a prior with Binomial distribution are shown in Fig.~\ref{fig:wdecay}. The results for other distributions are very similar and hence we do not include them here. Although we consider $400$ hypotheses to better analyse the effect of disruption, the results are similar with smaller, more realistic hypothesis spaces. Fig.~\ref{fig:entropy} shows that for larger disruptions (either due to lower value attention factor or higher frequency of its application), entropy is higher and therefore priors approach a uniform distribution and depart farther from the original Binomial distribution (the limit of the entropy is $\log _2400 = 8.64$ which corresponds to the uniform distribution). Also, Fig.~\ref{fig:unif} shows that as disruption increases (with fixed $r=0.8$ and increasing $t$), the output distribution approaches a uniform distribution. This implements the phenomenon of base-rate neglect as described in the last section. For large enough disruptions, the entropy reaches its maximum, and therefore the prior distribution becomes uniform, equivalent to complete base-rate neglect.

In sum, we can model base-rate neglect in the Bayesian framework by an attention module imposing weight disruption in our brain-like network, after prior and likelihood distributions are learned.  Note that weight disruption in our neural system could potentially simulate a range of biological and cognitive phenomena such as decline in attention or memory (partial use), deliberate neglect, or other ways of undermining the relevance of priors~\citep{Bar80}. The weight disruption effects could be all at once as when a prior network is not recalled or is judged irrelevant, or could take a long time reflecting the passage of time or disuse causing synaptic decay. Interference, the other main mechanism of memory decline, could likewise be examined within our neural-network system to implement and explain psychological interpretations of base-rate neglect.  
Our proposed neural network model contributes to the resolution of the discrepancy between demonstrated Bayesian successes and failures by modelling base-rate neglect as weight disruption in a connectionist network implementing Bayesian inference modulated by an attention module. 

\section{Discussion}
In a recent debate between critics~\citep{Bow12} and supporters~\citep{Gri12} of Bayesian models of cognition, probability matching becomes one of the points of discussion. Griffiths, et al. mention that probability matching phenomena have a ``key role in explorations of possible mechanisms for approximating Bayesian inference''~\citep[p. 420]{Gri12}. On the other hand, Bowers and Davis consider probability matching to be non--Bayesian, and propose an adaptive network that matches the posteriors as an alternative to the ``ad hoc and unparsimonious'' Bayesian account.  


We propose a framework which integrates these two seemingly opposing ideas. Instead of the network Bowers and Davis suggest to match the posterior probabilities,  we use probability modules to learn prior and likelihood distributions from realistic inputs in an autonomous and adaptive fashion. These distributions are later used in inferring MAP estimates. We show that our constructive neural network learns probability distributions naturally and in a psychologically realistic fashion through observable occurrence rates rather than being provided with explicit probabilities or stochastic units. We argue that probability modules with constructive neural networks provide a natural, autonomous way of introducing hypotheses and structures into Bayesian models. Recent demonstrations suggest that the fit of Bayes to human data depends crucially on assumptions of prior, and presumably likelihood, probability distributions~\citep{Mar13, Bow12}. Bayesian simulations would be less ad hoc if these probability distributions could be independently identified in human subjects rather than assumed by the modelers. The ability of neural networks to construct probability distributions from realistic observations of discrete events could likewise serve to constrain prior and likelihood distributions in simulations. Whether the full range of relevant hypotheses and structures can be constructed in this way deserves further exploration. The importance of our model is that, at the computational level, it is in accordance with  Bayesian accounts of cognition, and at the implementation level, it provides a psychologically--realistic account of learning and inference in humans. To the best of our knowledge, this is a novel way of integrating these opposing accounts.  Our proposed framework can be used in different models; for example, Dutta et al., used our modular framework for a post-stroke balance rehabilitation model~\citep{dutta14}.

In our framework, we use deterministic neural units. Deterministic units are of interest from a modelling perspective. It is important to see whether randomness and probabilistic representations can emerge as a property of a population of deterministic units rather than a built-in property of individual stochastic units. One can assume a model where a single stochastic unit produces outputs from a certain distribution, but that is engineered and not realistic or psychologically plausible. In this work, we consider very simple deterministic units and show that populations of these units can learn and represent probability distributions from realistic inputs and in an adaptive and autonomous fashion.

In this work, we introduce a max-product inference scheme to compute the MAP estimates; the outputs of prior and likelihood modules are combined in parallel to find the hypothesis with the highest posterior. This inference is fast and tractable for a number of reasons: (i) the prior and likelihoods are approximated with our SDCC modules and thus, their computation is fast and tractable (a single pass over the network); (ii) the product of prior and likelihoods is calculated in parallel neural circuitries and thus is not sensitive to the size of the problem; and (iii) since the often intractable denominator of Bayes’ rule is common among all hypotheses, we ignore that in computing the MAP estimate.

The question of the origins of Bayes' rule in biological learners remains unresolved. Future work on origins will undoubtedly examine the usual suspects of learning and evolution. Here we show that a modular map-product network can perform Bayesian inference. Our other in-progress work shows that simulated natural selection often favors a combination of individual learning and a Bayesian cultural ratchet in which a teacher's theory (represented as a distribution of posterior probabilities) serves as priors for a learner. Thus, both learning and evolution are still viable candidates, but many details of how they might act, alone or in concert, to produce Bayesian inference and learning  are yet to be worked out.

   In this introduction of our model, we deal with only a few Bayesian phenomena: learning probability distributions, probability matching, Bayes' rule, base-rate neglect, and relatively quick adapting to changing probabilities in the environment. There is a rapidly increasing number of other Bayesian phenomena that could provide interesting challenges to our neural model. So far, we are encouraged to see that the model can cover both Bayesian solutions and deviations from Bayes, promising a possible theoretical integration of disparate trends in the psychological literature. A number of apparent deviations from Bayesian optimality are listed elsewhere~\citep{Mar13}. In the cases we so far examined, deeper learning can convert deviations into something close to a Bayesian ideal, again suggesting the possibility of a unified account. 

With no doubt, Bayesian models provide powerful analytical tools to rigorously study deep questions of human cognition that have not been previously subject to formal analysis. These Bayesian ideas, providing computation-level models, are becoming prominent across a wide range of problems in cognitive science. The heuristic value of the Bayesian framework in providing insights into a wide range of psychological phenomena has been substantial, and in many cases unique.  Our neural implementation of probabilistic models addresses a number of recent challenges by allowing for the constrained construction of prior and likelihood distributions and greater generality in accounting for deviations from Bayesian ideals. As well, connectionist models offer an implementation-level framework for modeling mental phenomena in a more biologically plausible fashion.
 Providing network algorithms with the tools for doing Bayesian inference and learning could only enhance their power and utility. 
We present this work in the spirit of theoretical unification and mutual enhancement of these two approaches. We do not advocate replacement of one approach in favour of the other, but rather view the two approaches as being at different and complementary levels.  

\vspace*{-0.1 in}
\section*{Acknowledgement}
This work was supported by McGill Engineering Doctoral Award to MK, and an operating grant to TS from the Natural Sciences and Engineering Research Council of Canada. Mark Coates, Deniz \"Ustebay, and Peter Helfer contributed thoughtful comments on an earlier draft.

\section{References}
\bibliographystyle{model5-names}\biboptions{authoryear}
\bibliography{nipsbib}

\begin{thebibliography}{65}
\expandafter\ifx\csname natexlab\endcsname\relax\def\natexlab#1{#1}\fi
\providecommand{\url}[1]{\texttt{#1}}
\providecommand{\href}[2]{#2}
\providecommand{\path}[1]{#1}
\providecommand{\DOIprefix}{doi:}
\providecommand{\ArXivprefix}{arXiv:}
\providecommand{\URLprefix}{URL: }
\providecommand{\Pubmedprefix}{pmid:}
\providecommand{\doi}[1]{\href{http://dx.doi.org/#1}{\path{#1}}}
\providecommand{\Pubmed}[1]{\href{pmid:#1}{\path{#1}}}
\providecommand{\bibinfo}[2]{#2}
\ifx\xfnm\relax \def\xfnm[#1]{\unskip,\space#1}\fi
\bibitem[{Ackley et~al.(1985)Ackley, Hinton \& Sejnowski}]{Ack85}
\bibinfo{author}{Ackley, H.}, \bibinfo{author}{Hinton, G.}, \&
  \bibinfo{author}{Sejnowski, J.} (\bibinfo{year}{1985}).
\newblock \bibinfo{title}{A learning algorithm for {B}oltzmann machines}.
\newblock {\it \bibinfo{journal}{Cognitive Science}\/},  (pp.
  \bibinfo{pages}{147--169}).
\bibitem[{Baluja \& Fahlman(1994)}]{Bal94}
\bibinfo{author}{Baluja, S.}, \& \bibinfo{author}{Fahlman, S.~E.}
  (\bibinfo{year}{1994}).
\newblock {\it \bibinfo{title}{Reducing Network Depth in the
  Cascade-Correlation Learning Architecture}\/}.
\newblock \bibinfo{type}{Technical Report} Carnegie Mellon University, School
  of Computer Science.
\bibitem[{Bar-Hillel(1980)}]{Bar80}
\bibinfo{author}{Bar-Hillel, M.} (\bibinfo{year}{1980}).
\newblock \bibinfo{title}{The base-rate fallacy in probability judgments}.
\newblock {\it \bibinfo{journal}{Acta Psychologica}\/},  {\it
  \bibinfo{volume}{44}\/}, \bibinfo{pages}{211 -- 233}.
\bibitem[{Behrend \& Bitterman(1961)}]{Beh61}
\bibinfo{author}{Behrend, E.~R.}, \& \bibinfo{author}{Bitterman, M.}
  (\bibinfo{year}{1961}).
\newblock \bibinfo{title}{Probability-matching in the fish}.
\newblock {\it \bibinfo{journal}{The American Journal of Psychology}\/},  (pp.
  \bibinfo{pages}{542--551}).
\bibitem[{Bowers \& Davis(2012)}]{Bow12}
\bibinfo{author}{Bowers, J.~S.}, \& \bibinfo{author}{Davis, C.~J.}
  (\bibinfo{year}{2012}).
\newblock \bibinfo{title}{{Bayesian just-so stories in psychology and
  neuroscience.}}
\newblock {\it \bibinfo{journal}{Psychological Bulletin}\/},  {\it
  \bibinfo{volume}{138}\/}, \bibinfo{pages}{389--414}.
\bibitem[{Chater \& Manning(2006)}]{Cha06}
\bibinfo{author}{Chater, N.}, \& \bibinfo{author}{Manning, C.~D.}
  (\bibinfo{year}{2006}).
\newblock \bibinfo{title}{Probabilistic models of language processing and
  acquisition}.
\newblock {\it \bibinfo{journal}{Trends in Cognitive Sciences}\/},  {\it
  \bibinfo{volume}{10}\/}, \bibinfo{pages}{335--344}.
\bibitem[{Cover \& Thomas(2006)}]{Cov06}
\bibinfo{author}{Cover, T.~M.}, \& \bibinfo{author}{Thomas, J.~A.}
  (\bibinfo{year}{2006}).
\newblock {\it \bibinfo{title}{Elements of Information Theory ({W}iley Series
  in Telecommunications and Signal Processing)}\/}.
\newblock \bibinfo{publisher}{Wiley-Interscience}.
\bibitem[{Dawson et~al.(2009)Dawson, Dupuis, Spetch \& Kelly}]{Daw09}
\bibinfo{author}{Dawson, M.}, \bibinfo{author}{Dupuis, B.},
  \bibinfo{author}{Spetch, M.}, \& \bibinfo{author}{Kelly, D.}
  (\bibinfo{year}{2009}).
\newblock \bibinfo{title}{Simple artificial neural networks that match
  probability and exploit and explore when confronting a multiarmed bandit}.
\newblock {\it \bibinfo{journal}{IEEE Transactions on Neural Networks}\/},
  {\it \bibinfo{volume}{20}\/}, \bibinfo{pages}{1368--1371}.
\bibitem[{Dutta et~al.(2014)Dutta, Lahiri, Das, Nitsche \& Guiraud}]{dutta14}
\bibinfo{author}{Dutta, A.}, \bibinfo{author}{Lahiri, U.},
  \bibinfo{author}{Das, A.}, \bibinfo{author}{Nitsche, M.~A.}, \&
  \bibinfo{author}{Guiraud, D.} (\bibinfo{year}{2014}).
\newblock \bibinfo{title}{Post-stroke balance rehabilitation under multi-level
  electrotherapy: a conceptual review}.
\newblock {\it \bibinfo{journal}{{Frontiers in Neuroscience}}\/},  {\it
  \bibinfo{volume}{8}\/}.
\bibitem[{Eberhardt \& Danks(2011)}]{Ebe11}
\bibinfo{author}{Eberhardt, F.}, \& \bibinfo{author}{Danks, D.}
  (\bibinfo{year}{2011}).
\newblock \bibinfo{title}{{Confirmation in the cognitive sciences: The
  problematic case of Bayesian models}}.
\newblock {\it \bibinfo{journal}{Minds and Machines}\/},  {\it
  \bibinfo{volume}{21}\/}, \bibinfo{pages}{389--410}.
\bibitem[{Eddy(1982)}]{Edd82}
\bibinfo{author}{Eddy, D.~M.} (\bibinfo{year}{1982}).
\newblock \bibinfo{title}{Probabilistic reasoning in clinical medicine:
  problems and opportunities}.
\newblock In \bibinfo{editor}{D.~Kahneman}, \bibinfo{editor}{P.~Slovic}, \&
  \bibinfo{editor}{A.~Tversky} (Eds.), {\it \bibinfo{booktitle}{Judgment under
  uncertainty: Heuristics and biases}\/}.
\newblock \bibinfo{publisher}{Cambridge Univ. Press}.
\bibitem[{Evans et~al.(2002)Evans, Handley, Over \& Perham}]{Eva02}
\bibinfo{author}{Evans, J. S.~B.}, \bibinfo{author}{Handley, S.~J.},
  \bibinfo{author}{Over, D.~E.}, \& \bibinfo{author}{Perham, N.}
  (\bibinfo{year}{2002}).
\newblock \bibinfo{title}{Background beliefs in bayesian inference}.
\newblock {\it \bibinfo{journal}{Memory \& cognition}\/},  {\it
  \bibinfo{volume}{30}\/}, \bibinfo{pages}{179--190}.
\bibitem[{Fahlman(1988)}]{Fahl98}
\bibinfo{author}{Fahlman, S.~E.} (\bibinfo{year}{1988}).
\newblock \bibinfo{title}{Faster-learning variations on back-propagation: An
  empirical study}.
\newblock In {\it \bibinfo{booktitle}{Proc. of the Connectionist Models Summer
  School}\/} (pp. \bibinfo{pages}{38--51}).
\newblock \bibinfo{publisher}{Los Altos, CA: Morgan Kaufmann}.
\bibitem[{Fahlman \& Lebiere(1990)}]{Fah90}
\bibinfo{author}{Fahlman, S.~E.}, \& \bibinfo{author}{Lebiere, C.}
  (\bibinfo{year}{1990}).
\newblock \bibinfo{title}{The cascade-correlation learning architecture}.
\newblock In {\it \bibinfo{booktitle}{Advances in Neural Information Processing
  Systems 2}\/} (pp. \bibinfo{pages}{524--532}).
\newblock \bibinfo{publisher}{Loas Altos, CA: Morgan Kaufmann}.
\bibitem[{Gaissmaier \& Schooler(2008)}]{Gai08}
\bibinfo{author}{Gaissmaier, W.}, \& \bibinfo{author}{Schooler, L.~J.}
  (\bibinfo{year}{2008}).
\newblock \bibinfo{title}{The smart potential behind probability matching}.
\newblock {\it \bibinfo{journal}{Cognition}\/},  {\it \bibinfo{volume}{109}\/},
  \bibinfo{pages}{416--422}.
\bibitem[{Geman et~al.(1992)Geman, Bienenstock \& Doursat}]{Gem92}
\bibinfo{author}{Geman, S.}, \bibinfo{author}{Bienenstock, E.}, \&
  \bibinfo{author}{Doursat, R.} (\bibinfo{year}{1992}).
\newblock \bibinfo{title}{Neural networks and the bias/variance dilemma}.
\newblock {\it \bibinfo{journal}{Neural computation}\/},  {\it
  \bibinfo{volume}{4}\/}, \bibinfo{pages}{1--58}.
\bibitem[{Greggers \& Menzel(1993)}]{Gre93}
\bibinfo{author}{Greggers, U.}, \& \bibinfo{author}{Menzel, R.}
  (\bibinfo{year}{1993}).
\newblock \bibinfo{title}{Memory dynamics and foraging strategies of
  honeybees}.
\newblock {\it \bibinfo{journal}{Behavioral Ecology and Sociobiology}\/},  {\it
  \bibinfo{volume}{32}\/}, \bibinfo{pages}{17--29}.
\bibitem[{Griffiths et~al.(2012{\natexlab{a}})Griffiths, Austerweil \&
  Berthiaume}]{Grif12}
\bibinfo{author}{Griffiths, T.~L.}, \bibinfo{author}{Austerweil, J.~L.}, \&
  \bibinfo{author}{Berthiaume, V.~G.} (\bibinfo{year}{2012}{\natexlab{a}}).
\newblock \bibinfo{title}{Comparing the inductive biases of simple neural
  networks and bayesian models}.
\newblock In {\it \bibinfo{booktitle}{Proc. the 34th Annual Conf. of the Cog.
  Sci. Society}\/}.
\bibitem[{Griffiths et~al.(2012{\natexlab{b}})Griffiths, Chater, Norris \&
  Pouget}]{Gri12}
\bibinfo{author}{Griffiths, T.~L.}, \bibinfo{author}{Chater, N.},
  \bibinfo{author}{Norris, D.}, \& \bibinfo{author}{Pouget, A.}
  (\bibinfo{year}{2012}{\natexlab{b}}).
\newblock \bibinfo{title}{{ How the Bayesians got their beliefs (and what those
  beliefs actually are)}}.
\newblock {\it \bibinfo{journal}{Psychological Bulletin}\/},  {\it
  \bibinfo{volume}{138}\/}, \bibinfo{pages}{415--422}.
\bibitem[{Hampshire \& Pearlmutter(1990)}]{Ham90}
\bibinfo{author}{Hampshire, J.}, \& \bibinfo{author}{Pearlmutter, B.}
  (\bibinfo{year}{1990}).
\newblock \bibinfo{title}{Equivalence proofs for multi-layer perceptron
  classifiers and the {B}ayesian discriminant function}.
\newblock In {\it \bibinfo{booktitle}{Connectionist Models Summer School}\/}.
\bibitem[{Hardt et~al.(2013)Hardt, Nader \& Nadel}]{Har13}
\bibinfo{author}{Hardt, O.}, \bibinfo{author}{Nader, K.}, \&
  \bibinfo{author}{Nadel, L.} (\bibinfo{year}{2013}).
\newblock \bibinfo{title}{Decay happens: the role of active forgetting in
  memory}.
\newblock {\it \bibinfo{journal}{Trends in Cognitive Sciences}\/},  {\it
  \bibinfo{volume}{17}\/}, \bibinfo{pages}{111 -- 120}.
\bibitem[{Herrnstein(1961)}]{Her61}
\bibinfo{author}{Herrnstein, R.~J.} (\bibinfo{year}{1961}).
\newblock \bibinfo{title}{Relative and absolute strength of response as a
  function of frequency of reinforcement}.
\newblock {\it \bibinfo{journal}{Journal of the Experimental Analysis of
  Behaviour}\/},  {\it \bibinfo{volume}{4}\/}, \bibinfo{pages}{267--272}.
\bibitem[{Herrnstein(2000)}]{Her00}
\bibinfo{author}{Herrnstein, R.~J.} (\bibinfo{year}{2000}).
\newblock {\it \bibinfo{title}{The Matching Law: Papers on Psychology and
  Economics}\/}.
\newblock \bibinfo{address}{Cambridge, MA}: \bibinfo{publisher}{Harvard
  University Press}.
\bibitem[{Hinton(2010)}]{Hin10}
\bibinfo{author}{Hinton, G.} (\bibinfo{year}{2010}).
\newblock \bibinfo{title}{A practical guide to training restricted boltzmann
  machines}.
\newblock {\it \bibinfo{journal}{Momentum}\/},  {\it \bibinfo{volume}{9}\/},
  \bibinfo{pages}{926}.
\bibitem[{Hinton \& Osindero(2006)}]{Hin06}
\bibinfo{author}{Hinton, G.}, \& \bibinfo{author}{Osindero, S.}
  (\bibinfo{year}{2006}).
\newblock \bibinfo{title}{A fast learning algorithm for deep belief nets}.
\newblock {\it \bibinfo{journal}{Neural Computation}\/},  {\it
  \bibinfo{volume}{18}\/}, \bibinfo{pages}{1527 -- 1554}.
\bibitem[{Jaakkola et~al.(1996)Jaakkola, Saul \& Jordan}]{Jord2}
\bibinfo{author}{Jaakkola, T.~S.}, \bibinfo{author}{Saul, L.~K.}, \&
  \bibinfo{author}{Jordan, M.~I.} (\bibinfo{year}{1996}).
\newblock \bibinfo{title}{Fast learning by bounding likelihoods in sigmoid type
  belief networks}.
\newblock In {\it \bibinfo{booktitle}{Advances in Neural Information Processing
  Systems 22}\/}.
\bibitem[{James \& Koehler(2011)}]{Jam11}
\bibinfo{author}{James, G.}, \& \bibinfo{author}{Koehler, D.~J.}
  (\bibinfo{year}{2011}).
\newblock \bibinfo{title}{Banking on a bad bet probability matching in risky
  choice is linked to expectation generation}.
\newblock {\it \bibinfo{journal}{{Psychological Science}}\/},  {\it
  \bibinfo{volume}{22}\/}, \bibinfo{pages}{707--711}.
\bibitem[{Jones \& Love(2011)}]{Jon11}
\bibinfo{author}{Jones, M.}, \& \bibinfo{author}{Love, B.~C.}
  (\bibinfo{year}{2011}).
\newblock \bibinfo{title}{{Bayesian Fundamentalism or Enlightenment? On the
  explanatory status and theoretical contributions of Bayesian models of
  cognition}}.
\newblock {\it \bibinfo{journal}{Behavioral and Brain Sciences}\/},  {\it
  \bibinfo{volume}{34}\/}, \bibinfo{pages}{169--188}.
\bibitem[{Kahneman \& Tversky(1996)}]{Kah96}
\bibinfo{author}{Kahneman, D.}, \& \bibinfo{author}{Tversky, A.}
  (\bibinfo{year}{1996}).
\newblock \bibinfo{title}{On the reality of cognitive illusions}.
\newblock {\it \bibinfo{journal}{Psychological Review}\/},  {\it
  \bibinfo{volume}{103}\/}, \bibinfo{pages}{582 -- 591}.
\bibitem[{Kharratzadeh \& Shultz(2013)}]{Kha13}
\bibinfo{author}{Kharratzadeh, M.}, \& \bibinfo{author}{Shultz, T.}
  (\bibinfo{year}{2013}).
\newblock \bibinfo{title}{Neural-network modelling of {B}ayesian learning and
  inference}.
\newblock In {\it \bibinfo{booktitle}{{Proceedings of the 35th Annual Meeting
  of Cognitive Science}}\/} (pp. \bibinfo{pages}{2686--2691}).
\newblock \bibinfo{publisher}{Austin, TX: Cognitive Science Society}.
\bibitem[{Kirk \& Bitterman(1965)}]{Kir65}
\bibinfo{author}{Kirk, K.~L.}, \& \bibinfo{author}{Bitterman, M.}
  (\bibinfo{year}{1965}).
\newblock \bibinfo{title}{Probability-learning by the turtle}.
\newblock {\it \bibinfo{journal}{Science}\/},  {\it \bibinfo{volume}{148}\/},
  \bibinfo{pages}{1484--1485}.
\bibitem[{Koehler \& James(2009)}]{Koe09}
\bibinfo{author}{Koehler, D.~J.}, \& \bibinfo{author}{James, G.}
  (\bibinfo{year}{2009}).
\newblock \bibinfo{title}{Probability matching in choice under uncertainty:
  Intuition versus deliberation}.
\newblock {\it \bibinfo{journal}{Cognition}\/},  {\it \bibinfo{volume}{113}\/},
  \bibinfo{pages}{123--127}.
\bibitem[{Kruschke(2006)}]{Kru06}
\bibinfo{author}{Kruschke, J.~K.} (\bibinfo{year}{2006}).
\newblock \bibinfo{title}{Locally {B}ayesian learning with applications to
  retrospective revaluation and highlighting}.
\newblock {\it \bibinfo{journal}{Psychological Review}\/},  (pp.
  \bibinfo{pages}{677--699}).
\bibitem[{Lopez et~al.(1998)Lopez, Shanks, Almaraz \& Fernandez}]{Lop98}
\bibinfo{author}{Lopez, F.~J.}, \bibinfo{author}{Shanks, D.~R.},
  \bibinfo{author}{Almaraz, J.}, \& \bibinfo{author}{Fernandez, P.}
  (\bibinfo{year}{1998}).
\newblock \bibinfo{title}{Effects of trial order on contingency judgments: A
  comparison of associative and probabilistic contrast accounts.}
\newblock {\it \bibinfo{journal}{Journal of Experimental Psychology: Learning,
  Memory, and Cognition}\/},  {\it \bibinfo{volume}{24}\/},
  \bibinfo{pages}{672}.
\bibitem[{Ma et~al.(2006)Ma, Beck, Latham \& Pouget}]{Ma06}
\bibinfo{author}{Ma, W.~J.}, \bibinfo{author}{Beck, J.~M.},
  \bibinfo{author}{Latham, P.~E.}, \& \bibinfo{author}{Pouget, A.}
  (\bibinfo{year}{2006}).
\newblock \bibinfo{title}{{B}ayesian inference with probabilistic population
  codes}.
\newblock {\it \bibinfo{journal}{Nature Neuroscience}\/},  (pp.
  \bibinfo{pages}{1432 -- 1438}).
\bibitem[{Marcus \& Davis(2013)}]{Mar13}
\bibinfo{author}{Marcus, G.~F.}, \& \bibinfo{author}{Davis, E.}
  (\bibinfo{year}{2013}).
\newblock \bibinfo{title}{How robust are probabilistic models of higher-level
  cognition?}
\newblock {\it \bibinfo{journal}{Psychological Science}\/},  {\it
  \bibinfo{volume}{24}\/}, \bibinfo{pages}{2351--2360}.
\bibitem[{Marr(1982)}]{Mar82}
\bibinfo{author}{Marr, D.} (\bibinfo{year}{1982}).
\newblock {\it \bibinfo{title}{Vision}\/}.
\newblock \bibinfo{address}{San Francisco, CA}: \bibinfo{publisher}{W. H.
  Freeman}.
\bibitem[{McClelland(1998)}]{Mcc98}
\bibinfo{author}{McClelland, J.~L.} (\bibinfo{year}{1998}).
\newblock \bibinfo{title}{Connectionist models and bayesian inference}.
\newblock {\it \bibinfo{journal}{Rational models of cognition}\/},  (pp.
  \bibinfo{pages}{21--53}).
\bibitem[{McClelland et~al.(2014)McClelland, Mirman, Bolger \& Khaitan}]{Mcc14}
\bibinfo{author}{McClelland, J.~L.}, \bibinfo{author}{Mirman, D.},
  \bibinfo{author}{Bolger, D.~J.}, \& \bibinfo{author}{Khaitan, P.}
  (\bibinfo{year}{2014}).
\newblock \bibinfo{title}{Interactive activation and mutual constraint
  satisfaction in perception and cognition}.
\newblock {\it \bibinfo{journal}{Cognitive science}\/},  {\it
  \bibinfo{volume}{38}\/}, \bibinfo{pages}{1139--1189}.
\bibitem[{McClelland \& Rumelhart(1985)}]{Mccl85}
\bibinfo{author}{McClelland, J.~L.}, \& \bibinfo{author}{Rumelhart, D.~E.}
  (\bibinfo{year}{1985}).
\newblock \bibinfo{title}{Distributed memory and the representation of general
  and specific information}.
\newblock {\it \bibinfo{journal}{{Experimental Psychology: General}}\/},  {\it
  \bibinfo{volume}{114}\/}, \bibinfo{pages}{159--188}.
\bibitem[{Movellan \& McClelland(1993)}]{Mov93}
\bibinfo{author}{Movellan, J.}, \& \bibinfo{author}{McClelland, J.~L.}
  (\bibinfo{year}{1993}).
\newblock \bibinfo{title}{Learning continuous probability distributions with
  symmetric diffusion networks}.
\newblock {\it \bibinfo{journal}{Cognitive Science}\/},  {\it
  \bibinfo{volume}{17}\/}, \bibinfo{pages}{463--496}.
\bibitem[{Perfors et~al.(2011)Perfors, Tenenbaum, Griffiths \& Xu}]{Per11}
\bibinfo{author}{Perfors, A.}, \bibinfo{author}{Tenenbaum, J.~B.},
  \bibinfo{author}{Griffiths, T.~L.}, \& \bibinfo{author}{Xu, F.}
  (\bibinfo{year}{2011}).
\newblock \bibinfo{title}{A tutorial introduction to {B}ayesian models of
  cognitive development}.
\newblock {\it \bibinfo{journal}{Cognition}\/},  {\it \bibinfo{volume}{120}\/},
  \bibinfo{pages}{302 -- 321}.
\bibitem[{Prechelt(1998)}]{Pre98}
\bibinfo{author}{Prechelt, L.} (\bibinfo{year}{1998}).
\newblock \bibinfo{title}{Early stopping - but when?}
\newblock In \bibinfo{editor}{G.~Orr}, \& \bibinfo{editor}{K.-R. Muller}
  (Eds.), {\it \bibinfo{booktitle}{Neural Networks: Tricks of the Trade}\/}
  (pp. \bibinfo{pages}{55--69}).
\newblock \bibinfo{publisher}{Berlin: Springer} volume \bibinfo{volume}{1524}
  of {\it \bibinfo{series}{Lecture Notes in Computer Science}\/}.
\bibitem[{Prime \& Shultz(2011)}]{Pri11}
\bibinfo{author}{Prime, H.}, \& \bibinfo{author}{Shultz, T.~R.}
  (\bibinfo{year}{2011}).
\newblock \bibinfo{title}{Explicit {B}ayesian reasoning with frequencies,
  probabilities, and surprisals}.
\newblock In \bibinfo{editor}{C.~Hoelscher}, \bibinfo{editor}{T.~Shipley}, \&
  \bibinfo{editor}{L.~Carlson} (Eds.), {\it \bibinfo{booktitle}{Proceedings of
  33rd Annual Conference Cognitive Science Society}\/}.
\newblock \bibinfo{publisher}{{Boston, MA: Cognitive Science Society}}.
\bibitem[{Rumelhart et~al.(1995)Rumelhart, Durbin, Golden \& Chauvin}]{Rum95}
\bibinfo{author}{Rumelhart, D.~E.}, \bibinfo{author}{Durbin, R.},
  \bibinfo{author}{Golden, R.}, \& \bibinfo{author}{Chauvin, Y.}
  (\bibinfo{year}{1995}).
\newblock \bibinfo{title}{Backpropagation: The basic theory}.
\newblock In \bibinfo{editor}{Y.~Chauvin}, \& \bibinfo{editor}{D.~E. Rumelhart}
  (Eds.), {\it \bibinfo{booktitle}{Backpropagation: Theory, Arcitecture, and
  applications}\/} (pp. \bibinfo{pages}{1--34}).
\newblock \bibinfo{address}{Hillsdale, NJ, USA}.
\bibitem[{Saul et~al.(1996)Saul, Jaakkola \& Jordan}]{Jord1}
\bibinfo{author}{Saul, L.~K.}, \bibinfo{author}{Jaakkola, T.}, \&
  \bibinfo{author}{Jordan, M.~I.} (\bibinfo{year}{1996}).
\newblock \bibinfo{title}{Mean field theory for sigmoid belief networks}.
\newblock {\it \bibinfo{journal}{Journal of artificial intelligence
  research}\/},  {\it \bibinfo{volume}{4}\/}, \bibinfo{pages}{61--76}.
\bibitem[{Shanks(1990)}]{Shan90}
\bibinfo{author}{Shanks, D.~R.} (\bibinfo{year}{1990}).
\newblock \bibinfo{title}{Connectionism and the learning of probabilistic
  concepts}.
\newblock {\it \bibinfo{journal}{The Quarterly Journal of Experimental
  Psychology}\/},  {\it \bibinfo{volume}{42}\/}, \bibinfo{pages}{209--237}.
\bibitem[{Shanks(1991)}]{Shan91}
\bibinfo{author}{Shanks, D.~R.} (\bibinfo{year}{1991}).
\newblock \bibinfo{title}{A connectionist account of base-rate biases in
  categorization}.
\newblock {\it \bibinfo{journal}{Connection Science}\/},  {\it
  \bibinfo{volume}{3}\/}, \bibinfo{pages}{143--162}.
\bibitem[{Shanks et~al.(2002)Shanks, Tunney \& McCarthy}]{Sha02}
\bibinfo{author}{Shanks, D.~R.}, \bibinfo{author}{Tunney, R.~J.}, \&
  \bibinfo{author}{McCarthy, J.~D.} (\bibinfo{year}{2002}).
\newblock \bibinfo{title}{A re-examination of probability matching and rational
  choice}.
\newblock {\it \bibinfo{journal}{Journal of Behavioral Decision Making}\/},
  {\it \bibinfo{volume}{15}\/}, \bibinfo{pages}{233--250}.
\bibitem[{Shultz(2003)}]{Shu13}
\bibinfo{author}{Shultz, T.} (\bibinfo{year}{2003}).
\newblock {\it \bibinfo{title}{Computational Developmental Psychology}\/}.
\newblock \bibinfo{address}{Cambridge, MA}: \bibinfo{publisher}{MIT Press}.
\bibitem[{Shultz(2007)}]{Shu07}
\bibinfo{author}{Shultz, T.} (\bibinfo{year}{2007}).
\newblock \bibinfo{title}{{The Bayesian revolution approaches psychological
  development}}.
\newblock {\it \bibinfo{journal}{Developmental Science}\/},  {\it
  \bibinfo{volume}{10}\/}, \bibinfo{pages}{357--364}.
\bibitem[{Shultz(2012)}]{Shu12-2}
\bibinfo{author}{Shultz, T.} (\bibinfo{year}{2012}).
\newblock \bibinfo{title}{A constructive neural-network approach to modeling
  psychological development}.
\newblock {\it \bibinfo{journal}{Cognitive Development}\/},  {\it
  \bibinfo{volume}{27}\/}, \bibinfo{pages}{383--400}.
\bibitem[{Shultz(2013)}]{Shu13-2}
\bibinfo{author}{Shultz, T.} (\bibinfo{year}{2013}).
\newblock \bibinfo{title}{Computational models of developmental psychology}.
\newblock In \bibinfo{editor}{P.~D. Zelazo} (Ed.), {\it
  \bibinfo{booktitle}{Oxford Handbook of developmental Psychology, Vol. 1: Body
  and mind}\/}.
\newblock \bibinfo{address}{Newyork}: \bibinfo{publisher}{Oxford University
  Press}.
\bibitem[{Shultz et~al.(2012)Shultz, Doty \& Dandurand}]{Shu12n}
\bibinfo{author}{Shultz, T.}, \bibinfo{author}{Doty, E.}, \&
  \bibinfo{author}{Dandurand, F.} (\bibinfo{year}{2012}).
\newblock \bibinfo{title}{Knowing when to abandon unproductive learning}.
\newblock In {\it \bibinfo{booktitle}{{Proceedings of the 34th Annual
  Conference of the Cognitive Science Society}}\/} (pp.
  \bibinfo{pages}{2327--2332}).
\newblock \bibinfo{address}{Austin, TX: Cognitive Science Society}.
\bibitem[{Tenenbaum et~al.(2006)Tenenbaum, Kemp \& Shafto}]{Ten06}
\bibinfo{author}{Tenenbaum, J.~B.}, \bibinfo{author}{Kemp, C.}, \&
  \bibinfo{author}{Shafto, P.} (\bibinfo{year}{2006}).
\newblock \bibinfo{title}{Theory-based {B}ayesian models of inductive learning
  and reasoning}.
\newblock {\it \bibinfo{journal}{Trends in Cognitive Sciences}\/},  {\it
  \bibinfo{volume}{10}\/}, \bibinfo{pages}{309--318}.
\bibitem[{Unturbe \& Corominas(2007)}]{Unt07}
\bibinfo{author}{Unturbe, J.}, \& \bibinfo{author}{Corominas, J.}
  (\bibinfo{year}{2007}).
\newblock \bibinfo{title}{Probability matching involves rule-generating
  ability: A neuropsychological mechanism dealing with probabilities.}
\newblock {\it \bibinfo{journal}{Neuropsychology}\/},  {\it
  \bibinfo{volume}{21}\/}, \bibinfo{pages}{621}.
\bibitem[{Vulkan(2000)}]{Vul98}
\bibinfo{author}{Vulkan, N.} (\bibinfo{year}{2000}).
\newblock \bibinfo{title}{An economist's perspective on probability matching}.
\newblock {\it \bibinfo{journal}{Journal of Economic Surveys}\/},  {\it
  \bibinfo{volume}{14}\/}, \bibinfo{pages}{101--118}.
\bibitem[{Wang et~al.(1993)Wang, Venkatesh \& Judd}]{Wan93}
\bibinfo{author}{Wang, C.}, \bibinfo{author}{Venkatesh, S.~S.}, \&
  \bibinfo{author}{Judd, J.~S.} (\bibinfo{year}{1993}).
\newblock \bibinfo{title}{Optimal stopping and effective machine complexity in
  learning}.
\newblock In {\it \bibinfo{booktitle}{Advances in Neural Information Processing
  Systems 6}\/} (pp. \bibinfo{pages}{303--310}).
\newblock \bibinfo{publisher}{Morgan Kaufmann}.
\bibitem[{West \& Stanovich(2003)}]{Wes03}
\bibinfo{author}{West, R.~F.}, \& \bibinfo{author}{Stanovich, K.~E.}
  (\bibinfo{year}{2003}).
\newblock \bibinfo{title}{Is probability matching smart? associations between
  probabilistic choices and cognitive ability}.
\newblock {\it \bibinfo{journal}{Memory \& Cognition}\/},  {\it
  \bibinfo{volume}{31}\/}, \bibinfo{pages}{243--251}.
\bibitem[{White(1989)}]{Whi89}
\bibinfo{author}{White, H.} (\bibinfo{year}{1989}).
\newblock \bibinfo{title}{Learning in artificial neural networks: A statistical
  perspective}.
\newblock {\it \bibinfo{journal}{Neural computation}\/},  {\it
  \bibinfo{volume}{1}\/}, \bibinfo{pages}{425--464}.
\bibitem[{Wolford et~al.(2000)Wolford, Miller \& Gazzaniga}]{Wol00}
\bibinfo{author}{Wolford, G.}, \bibinfo{author}{Miller, M.~B.}, \&
  \bibinfo{author}{Gazzaniga, M.} (\bibinfo{year}{2000}).
\newblock \bibinfo{title}{The left hemisphere's role in hypothesis formation.}
\newblock {\it \bibinfo{journal}{The Journal of Neuroscience}\/},  {\it
  \bibinfo{volume}{20}\/}, \bibinfo{pages}{1--4}.
\bibitem[{Wolford et~al.(2004)Wolford, Newman, Miller \& Wig}]{Wol04}
\bibinfo{author}{Wolford, G.}, \bibinfo{author}{Newman, S.~E.},
  \bibinfo{author}{Miller, M.~B.}, \& \bibinfo{author}{Wig, G.~S.}
  (\bibinfo{year}{2004}).
\newblock \bibinfo{title}{Searching for patterns in random sequences.}
\newblock {\it \bibinfo{journal}{Canadian Journal of Experimental
  Psychology}\/},  {\it \bibinfo{volume}{58}\/}, \bibinfo{pages}{221}.
\bibitem[{Wozny et~al.(2010)Wozny, Beierholm \& Shams}]{Woz10}
\bibinfo{author}{Wozny, D.~R.}, \bibinfo{author}{Beierholm, U.~R.}, \&
  \bibinfo{author}{Shams, L.} (\bibinfo{year}{2010}).
\newblock \bibinfo{title}{Probability matching as a computational strategy used
  in perception}.
\newblock {\it \bibinfo{journal}{PLoS computational biology}\/},  {\it
  \bibinfo{volume}{6}\/}, \bibinfo{pages}{e1000871}.
\bibitem[{Yellott~Jr(1969)}]{Yel69}
\bibinfo{author}{Yellott~Jr, J.~I.} (\bibinfo{year}{1969}).
\newblock \bibinfo{title}{Probability learning with noncontingent success}.
\newblock {\it \bibinfo{journal}{Journal of {M}athematical {P}sychology}\/},
  {\it \bibinfo{volume}{6}\/}, \bibinfo{pages}{541--575}.
\bibitem[{Yuille \& Kersten(2006)}]{Yui06}
\bibinfo{author}{Yuille, A.}, \& \bibinfo{author}{Kersten, D.}
  (\bibinfo{year}{2006}).
\newblock \bibinfo{title}{Vision as {B}ayesian inference: analysis by
  synthesis?}
\newblock {\it \bibinfo{journal}{Trends in Cognitive Sciences}\/},  {\it
  \bibinfo{volume}{10}\/}, \bibinfo{pages}{301--308}.

\end{thebibliography}

\end{document}